\pdfoutput=1
\documentclass[11pt]{article}

\PassOptionsToPackage{table}{xcolor}
\usepackage[table]{xcolor} 
\usepackage[preprint]{acl}

\usepackage{times}
\usepackage{latexsym}
\usepackage{amsmath,amssymb}
\usepackage{booktabs}
\usepackage{multirow}
\usepackage{graphicx}
\usepackage{xspace}
\usepackage{enumitem}
\newcommand{\cc}{\cellcolor{gray!20}} 
\usepackage{pifont}
\usepackage{amssymb}
\newcommand{\cmark}{\ding{51}}%
\newcommand{\xmark}{\ding{55}}%

\usepackage[T1]{fontenc}

\usepackage[utf8]{inputenc}

\usepackage{microtype}

\usepackage{inconsolata}

\usepackage{graphicx}
\usepackage{textcomp}
\usepackage{float}

\usepackage{multicol}
\usepackage{wrapfig}
\def \name{\textsc{CoPe}\xspace}

\title{Personalized LLM Decoding via Contrasting Personal Preference}
\author{
Hyungjune Bu\thanks{Equal contribution (listed in alphabetical order).}\textsuperscript{,1},~~
Chanjoo Jung\footnotemark[1]\textsuperscript{,1},~~
Minjae Kang\textsuperscript{2},~~
Jaehyung Kim\textsuperscript{1} \\
\textsuperscript{1}Yonsei University, 
\textsuperscript{2}Opt-AI Inc. \\
\texttt{cleverscent@yonsei.ac.kr},~~ \texttt{chanjoo0427@yonsei.ac.kr}
}

\usepackage[absolute,overlay]{textpos}
\usepackage{wrapfig}
\usepackage{float}
\usepackage{subcaption}  

\usepackage{listings}

\usepackage{enumitem}
\usepackage{pifont}

\begin{document}

\maketitle

\begingroup
\renewcommand\thefootnote{*}
\endgroup

\begin{abstract}
As large language models (LLMs) are progressively deployed in various real-world applications, personalization of LLMs has become increasingly important. 
While various approaches to LLM personalization such as prompt-based and training-based methods have been actively explored, the development of effective decoding-time algorithms remains largely overlooked, despite their demonstrated potential.
In this paper, we propose \name{} (\textbf{Co}ntrasting \textbf{Pe}rsonal Preference), a novel decoding-time approach applied after performing parameter-efficient fine-tuning (PEFT) on user-specific data. 
Our core idea is to leverage reward-guided decoding specifically for personalization by maximizing each user's implicit reward signal. 
We evaluate \name{} across five open-ended personalized text generation tasks. 
Our empirical results demonstrate that \name{} achieves strong performance, improving personalization by an average of 10.57\% in ROUGE-L,without relying on external reward models or additional training procedures.\footnote{Code is available at \url{https://github.com/cleverscent/CoPe}.}
\end{abstract}

\section{Introduction}

Personalization of Large Language Models (LLMs) \citep{achiam2023gpt, team2023gemini, claude3, touvron2023llama2openfoundation}, which refers to the process of aligning model outputs with individual user preferences, has received growing attention as LLMs are increasingly deployed in real-world applications such as writing assistants \citep{mysore2024pearlpersonalizinglargelanguage}, content recommendation \citep{zhang2024sparpersonalizedcontentbasedrecommendation}, and review generation \citep{peng2024reviewllmharnessinglargelanguage}. 
Prompt-based personalization \citep{santurkar2023whose, hwang2023aligning}, which augments a user query by retrieving prior interactions or constructing a summarized user profile, is arguably considered as one of the most straightforward approaches.
However, its effectiveness is often limited by the absence of direct learning from user data.
In contrast, training-based personalization \citep{zhao2024group, kim2025few} captures user preferences more effectively by updating model parameters, but it also suffers from challenges such as catastrophic forgetting and increased computational costs.
To mitigate these limitations, recent works such as One PEFT per User \citep{tan2024democratizing} have demonstrated that lightweight parameter-efficient fine-tuning (PEFT) offers a viable solution for personalizing LLMs \citep{tan2025democratizinglargelanguagemodels, zhang2025personalizedllmresponsegeneration, kim2025personalized}. 
Unlike prior works mentioned above, we turn to a new perspective for effective LLM personalization.\looseness=-1
\begin{figure}[t]
  \centering
  \hspace*{-1em}  
  \includegraphics[width=1.0\linewidth]{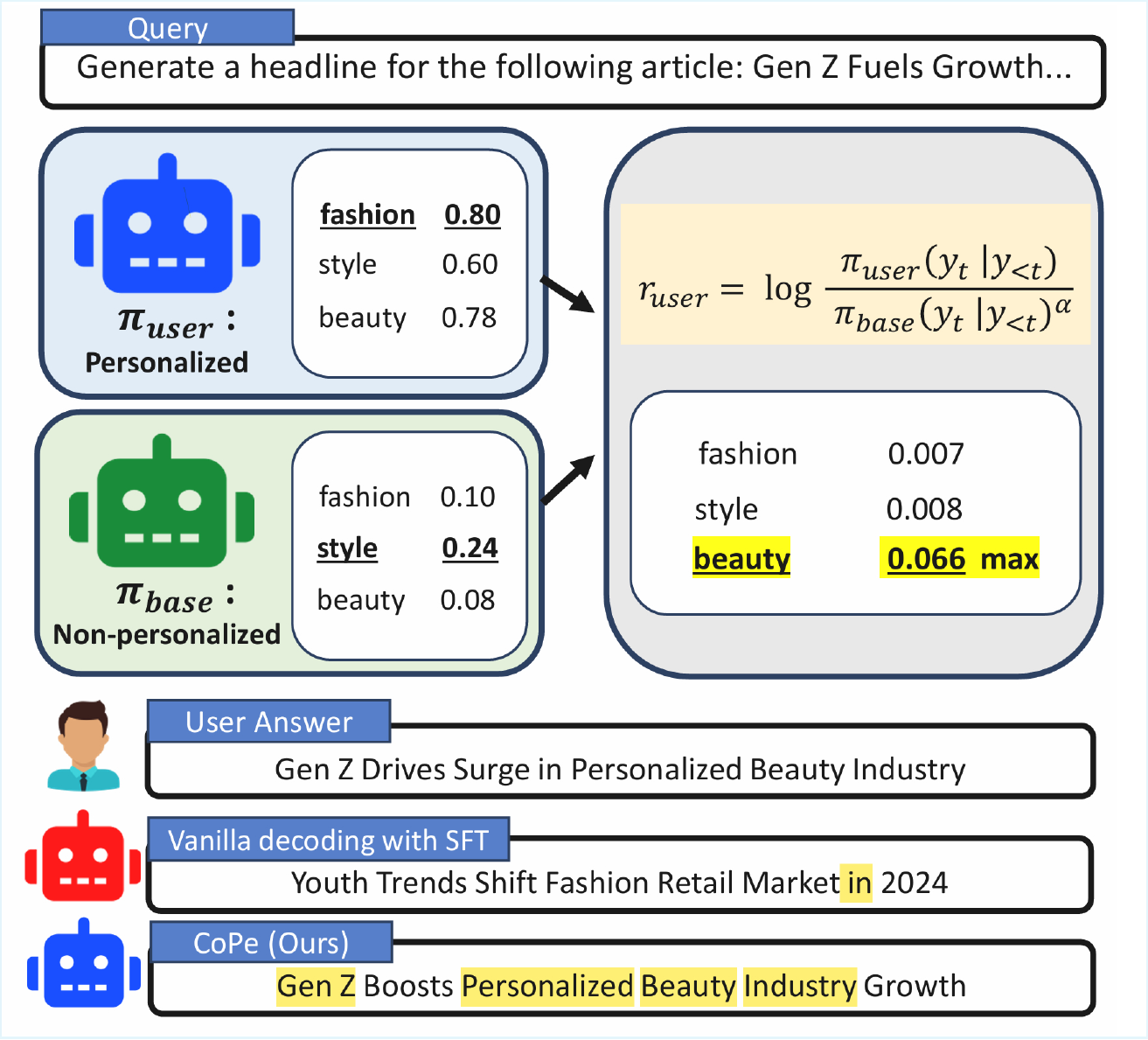}

  \caption{\textbf{Implicit reward maximization via contrastive preference}. Under an implicit reward model that leverages the interaction between a personalized and a non-personalized generic model, generated texts better align with user preferences. The highlighted text marks words that overlaps with the gold answer.}
\label{fig:cope_overview}
\end{figure}

\begin{figure*}[t]
    \centering
    \hspace*{0cm} 
    \includegraphics[
        width=1.0\textwidth,
        trim=0 0 0 0,
        clip
    ]{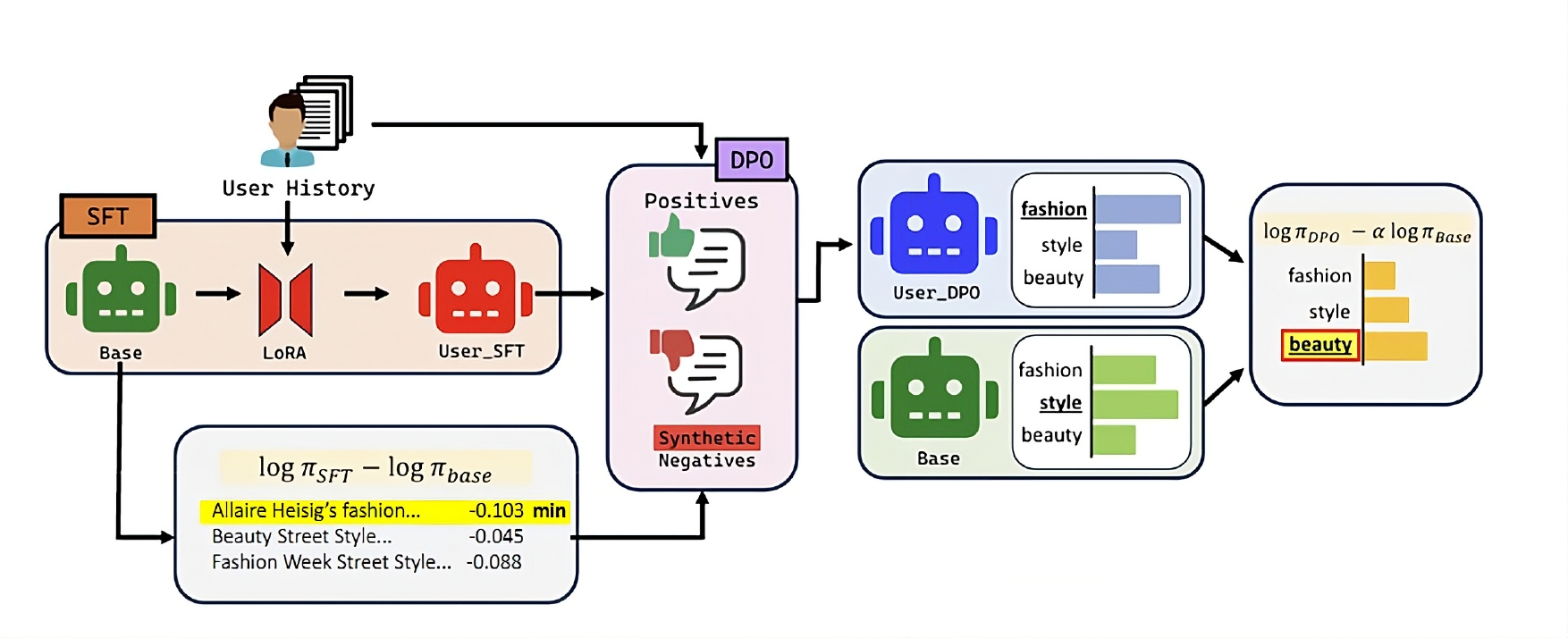}

    \caption{\textbf{Illustration of \name{}} ({\normalsize\textbf{Co}}ntrasting Preference for {\normalsize\textbf{Pe}}rsonalized LLM Decoding). The training pipeline (left) builds an expert user model via Direct Preference Optimization (DPO) with synthetic negatives. The reward-guided decoding method (right) contrasts this user model with a base model at the token level, maximizing implicit user reward during both training and decoding for improved personalization.}

    \label{fig:method}
\end{figure*}

In this work, we introduce \name{} (\textbf{Co}ntrasting \textbf{Pe}rsonal preference), a new paradigm for LLM personalization that operates at the decoding stage, applied after PEFT on user-specific data. 
At a high level, \name{} is a form of \textit{reward-guided decoding} \citep{deng2023reward,khanov2024args,lightman2024let}, an approach that effectively steers LLM outputs toward desired properties (\textit{e.g.}, improved reasoning) by maximizing a reward function, adapted specifically for personalizing LLMs across varying contexts and user goals.

Unlike conventional reward-guided decoding methods, \name{} does not require an external reward model to estimate rewards. 
Instead, it leverages the implicit user reward signal, which can be efficiently approximated using the likelihoods from both the PEFT-tuned model and the original base model. 
Building on our key insight which connects this implicit reward to the objective of contrastive decoding \citep{Li2023contrastive},we can implement the proposed \name{} easily (see overview in Figure \ref{fig:method}).

In addition, we further enhance PEFT for LLM personalization by encouraging the model to better capture the implicit user reward. 
The core idea is to contrast implicit rewards between a \textit{positive} response (provided by the user) and a \textit{negative} response (unlikely to be from the user, \textit{e.g.}, from other users), using Direct Preference Optimization (DPO) \citep{rafailov2023dpo}. 
To avoid the practical and privacy challenges of relying on data from other users, we synthesize negative responses by generating outputs with low implicit rewards via Best-of-N sampling \citep{gui2024bonbon}.
This training method not only improves the effectiveness of PEFT, but also enhances the performance of our proposed reward-guided decoding by enabling more accurate modeling of the implicit user reward. 
An overview of the pipeline is shown in Figure \ref{fig:method}.

We demonstrate the effectiveness of \name{} with experiments in five different personalized open-ended text generation tasks from Language Model Personalization (LaMP) \citep{salemi2024lamp} and LongLaMP \citep{kumar2024longlamp} benchmarks. 
Specifically, \name{} achieves an average relative improvement of 10.57\% in ROUGE-L across all tasks, compared to the task-finetuned model. 
Notably, \name{} also outperforms a simply personalized model that lacks the contrastive mechanism, with an average ROUGE-L gain of 5.67\% across tasks. 
Furthermore, the effectiveness of \name{} is well-generalized across different scales and types of state-of-the-art LLMs. 
Our robust experimental results show that the implicit reward maximization of \name{} further enhances alignment with individual user preferences. 
Together, these findings highlight \name{} as a promising approach for scalable and effective LLM personalization.

\section{Related Works}

\paragraph{LLM personalization.}
Given the diversity of user goals and preferences, various approaches to personalization of LLM have been explored. 
One common strategy is prompt-based personalization, wherein techniques such as retrieval-augmented generation (RAG) \citep{lewis2021retrievalaugmentedgenerationknowledgeintensivenlp} and prompt-augmented generation (PAG) \citep{richardson2023integratingsummarizationretrievalenhanced} dynamically inject user-specific context into each prompt at inference.
However, these methods lack parametric memory and rely entirely on prompt construction, making them vulnerable to context length limitations and insufficient grounding. 
On the other hand, training-based personalization methods, which fine-tune the model on user-specific data, have demonstrated superior performance in capturing user preferences compared to prompting-based approaches \citep{zhao2024group,zhuang2024hydra}. 
Nevertheless, even these methods face several limitations. Firstly, these methods are computationally intensive, as they involve modifying model parameters. 
In fact, in the worst case, frequent retraining may be necessary to reflect evolving user preferences \citep{madotto-etal-2021-continual}. 
Moreover, these methods are susceptible to catastrophic forgetting, a phenomenon in which adapting to new user data can lead the model to forget previously learned preferences or even general knowledge \citep{MCCLOSKEY1989109, dautume2019episodicmemorylifelonglanguage}.\looseness=-1

A recent and practical method to address these limitations is the utilization of lightweight parameter-efficient fine-tuning (PEFT), which offers an effective and scalable approach to personalizing LLMs \citep{zhang2024sparpersonalizedcontentbasedrecommendation, zhang2025personalizationlargelanguagemodels}. 
Meanwhile, personalization at the decoding stage remains largely unexplored in existing methods. 
Motivated by this gap, we aim to address the aforementioned limitations through a decoding-based approach to personalization.\looseness=-1

\paragraph{LLM decoding.}
Various decoding strategies have been explored and applied in LLMs to boost their performance. 
For instance, contrastive decoding has demonstrated strong effectiveness not only in open-ended text generation \citep{Li2023contrastive}, but also in reasoning \citep{obrien2023contrastivedecodingimprovesreasoning}, retrieval-augmented generation (RAG) \citep{shi2023trustingevidencehallucinatecontextaware}, and even multi-modal generation \citep{leng2023mitigatingobjecthallucinationslarge}. On the other hand, reward-guided decoding has emerged as another promising approach, aiming to improve alignment and reasoning capabilities directly at the decoding stage, without additional model training. 
To further explain, reward-guided decoding guides the generation process using reward signals, offering a lightweight yet effective alternative for steering outputs toward desired behaviors \citep{deng2023reward, lightman2024let}. 
In fact, adaptive reward shaping, as proposed by \citet{khanov2024args}, has also been shown to improve sample efficiency during decoding. 
Despite the growing interest in both decoding strategies and personalization, there is no prior work that effectively leverages decoding methods for personalization due to the challenge of modeling separate rewards for each user.
In this aspect, we propose the first guided decoding approach for personalization that does not require any external reward models. 
Specifically, our method can be easily implemented using contrastive decoding, thereby enabling more practical and scalable deployment in real world settings.\looseness=-1

\paragraph{Preference learning.}
Preference learning is an approach that ensures alignment with human or task-specific preferences by leveraging relative feedback between outputs, rather than relying on absolute labels. 
One traditional approach to preference learning is Reinforcement Learning from Human Feedback (RLHF) \citep{ouyang2022traininglanguagemodelsfollow}, which involves fitting a reward model based on human-labeled comparisons and optimizing model policies through reinforcement learning.
However, RLHF often requires complex and costly training procedures. 
To address this limitation, recent methods such as Direct Preference Optimization (DPO) \citep{rafailov2023dpo} simplify the process by directly fine-tuning models through binary classification between preferred and dispreferred outputs. 

Building on these advances, we propose a personalized fine-tuning method that integrates preference learning by treating user profile responses as positive examples and non personalized outputs as negative examples. 
This training formulation supports contrastive decoding, due to the fact that maximization of implicit user reward is plausible both in the training and decoding section.
In other words, this conceptual alignment between preference learning and contrastive decoding ensures consistency between training and inference, enabling more effective personalization without external reward models or additional training procedures.

\section{\name{}: Contrasting Preference for Personalized LLM Decoding}

In this section, we present our new decoding framework for LLM personalization by \textbf{Co}ntrasting \textbf{Pe}rsonal preference (\name{}). 
Our key idea is incorporating \textit{implicit reward signals} for user preference to guide both training and inference. 
We first present our problem setup in Section~\ref{sec2.1:pre}. 
Next, we present the proposed decoding scheme, \name{}, in Section \ref{sec3.1:decode}. 
Lastly, in Section \ref{sec3.2:dpo}, we present our training scheme to further improve PEFT for personalization, by explicitly maximizing user reward based on the synthetic negative response.

\subsection{Preliminary}\label{sec2.1:pre}
Let us first assume that we have the historical interaction data $H_{\tt user} = \{(x^i, y^i)\}_{i=1}^{N}$ for a target user.
Then, for a given input query $x$, the goal of LLM personalization is to generate a personalized output $y$ from LLM $\pi$ that aligns with the user's preferences and behaviors exhibited in $H_{\tt user}$.
A representative approach for LLM personalization is to adapt a generic pre-trained LLM $\pi_{\tt base}$ using parameter-efficient fine-tuning (PEFT) techniques, such as LoRA \citep{hu2021loralowrankadaptationlarge}.

Formally, let $\Delta_{\tt user}$ denote the user-specific PEFT module.\footnote{In this work, we only consider LoRA.} 
The personalized model is then defined as $\pi_{\tt user} = \pi_{\tt base} + \Delta_{\tt user}$, such that only $\Delta_{\tt user}$ is optimized using the user's data $H_{\tt user}$.
For example, \citet{tan2024democratizing} optimizes $\Delta_{\tt user}$ on $H_{\tt user}$ via conventional supervised fine-tuning (SFT) that minimizes cross-entropy between the output of $\pi_{\tt user}(x^i)$ and ground-truth label $y^i$.
After optimizing $\Delta_{\tt user}$, $\pi_{\tt user}$ is expected to generate the responses that align with the user's preferences.  

\subsection{Optimizing personal preference via contrastive decoding with PEFT}\label{sec3.1:decode}

Assume that we have access to a base model $\pi_{\tt base}$ and a personalized model $\pi_{\tt user}$.
Then, to generate a response $y$ that better aligns with the user's preferences for a given test query $x$, \name{} adopts a reward-guided decoding strategy that contrasts the token-level likelihoods under these two models.

Let ${y}_{<t} = (y_1, \dots, y_{t-1})$ denote the partial output sequence at decoding step $t$. 
Then, following \citet{Li2023contrastive}, we first define a plausibility-constrained candidate set of next tokens as:
\begin{equation}
    \mathcal{V}_{\text{head}}^t = \big\{y_t \in \mathcal{V} \,\big|\, \pi_{\tt user}(y_t \mid y_{<t}) \ge \tau_t\},
\end{equation}

\noindent where $\tau_t:= \tau \cdot \max_{w \in \mathcal{V}} \pi_{\tt user}(w \mid y_{<t})$ is an adaptive threshold determined by a hyperparameter $\tau \in [0, 1]$ and $\mathcal{V}$ denotes the vocabulary for $\pi_{\tt user}$. 
For each candidate token $y_t \in \mathcal{V}_{\text{head}}^t$, we compute an implicit user reward by contrasting its likelihoods under the personalized and base models:
\begin{equation}
    r_{\tt {user}}(y_t) = \log \frac{\pi_{\tt {user}}(y_t \mid y_{<t})}{\pi_{\tt {base}}(y_t \mid y_{<t})^{\alpha}},\label{eq:reward}
\end{equation}
where $\alpha \ge 0$ is a contrastive weight hyperparameter.
This reward encourages the selection of tokens that are strongly preferred by the personalized model while being penalized under the base model, yields the outputs that are both user-aligned and distinctive. 
Finally, the next token $y_t^{*}$ is selected which maximizes the implicit user reward:
\begin{equation}
    y_t^* = \arg\max_{y_t \in \mathcal{V}_{\text{head}}^t} r_{\tt {user}}(y_t).\label{eq:cope_decode}
\end{equation}

\paragraph{Rationale behind implicit user reward.} 
Here, we present the theoretical intuition behind our proposed implicit user reward $r_{\tt user}$ (Eq.~\ref{eq:reward}). 
To this end, we revisit the concept of \textit{implicit reward} introduced in DPO \citep{rafailov2023dpo}, which has been widely adopted in the LLM alignment literature \citep{chen2025bootstrapping, kim2025spread, cui2025process}.
Specifically, \citet{rafailov2023dpo} show that the reward function $r$, which captures human preferences, can be approximated under the RLHF framework \citep{ouyang2022traininglanguagemodelsfollow} as the log-likelihood ratio between the optimal (aligned) LLM policy $\pi_r$ and a reference policy $\pi_{\tt ref}$:
\begin{equation}
    r(y)\approx\beta\cdot\log\frac{\pi_r(y)}{\pi_{\tt ref}(y)},\label{eq:dpo_implicit}
\end{equation}
where $\beta$ is a hyperparameter controlling the strength of KL regularization in RLHF.\footnote{While $y$ is generated for input $x$, we omit this in Eq.~\ref{eq:dpo_implicit} for the simplicity.}
This derivation of implicit reward enables reward modeling without an explicit reward model using only the relative likelihoods under two LLM policies, and yields a much more efficient preference learning algorithm, called DPO (see details in Appendix \ref{app:rlhf_dpo}). 

In our setting, however, the personalized model $\pi_{\tt user}$ is not trained with explicit KL regularization, as in standard RLHF.
Nevertheless, we argue that the PEFT used for training $\pi_{\tt user}$ implicitly imposes a similar constraint. 
For example, in LoRA \citep{hu2021loralowrankadaptationlarge}, only the newly introduced low-rank matrices are updated, while the original model parameters remain fixed. 
This architectural constraint implicitly regularizes the updated model, preventing it from deviating significantly from the base model.
As a result, the personalized model $\pi_{\tt user}$ trained via PEFT remains close to the base model $\pi_{\tt base}$, and the log-likelihood ratio between them can serve as a valid proxy for an implicit reward signal—namely, $r_{\tt user}$.
We further empirically validate that these log-likelihood ratios (Eq.~\ref{eq:reward}) encode meaningful personalized signals through detailed analyses and results in Appendix \ref{app:implicit-reward-validation}. 

Interestingly, we note that this formulation, based on the ratio of log-likelihoods between two models, also appears in contrastive decoding \citep{Li2023contrastive}. 
In this sense, our insight reveals a novel connection between two popular decoding paradigms, contrastive decoding and reward-guided decoding.
Following \citet{Li2023contrastive}, we additionally introduce a hyperparameter $\alpha$ to control the strength of contrastive adjustment during decoding and further enhance personalization.

\subsection{Aligning PEFT to user preference via DPO with synthetic negative response}\label{sec3.2:dpo}

While \name{} effectively maximizes the implicit user reward during decoding with the personalized model $\pi_{\tt user}$, its performance can be further improved by explicitly aligning $\pi_{\tt user}$ with the user's  individual preferences during training.

One natural approach is to apply preference learning algorithms such as RLHF or DPO. 
However, a key practical challenge is a lack of negative examples ((\textit{i.e.}, responses unlikely to come from the user) in the user dataset $H_{\tt user}$.
To address this, we propose a simple yet effective approach that synthesizes negative examples leveraging the implicit user reward $r_{\tt user}$.
Specifically, for each train query $x^i \in H_{\tt user}$, we sample $K$ candidate responses $\{\widetilde{y}^{i,1}, \dots, \widetilde{y}^{i,K}\}$ from the generic base model $\pi_{\tt base}$.
Among these, we select the response with the lowest implicit user reward, \textit{i.e.}, the one that is most unlikely from the user:
\begin{equation}
    \widetilde{y}^{i,*} = \arg \min_{y \in 
 \{\widetilde{y}^{i,1}, \dots, \widetilde{y}^{i,K}\}} \sum_{t} r_{\tt user}({y_t}),\label{eq:reward_neg}
\end{equation}
\noindent where the contrastive weight $\alpha$ is set to $1$.
Then, we construct a preference dataset $\mathcal{D}_{\text{pref}}:=\{(x^i,y^i_{\tt pos},y^i_{\tt neg})\}_{i=1}^{N}$ where $(x^i,y^i_{\tt pos})$ from $H_u$, \textit{i.e.}, $y^i_{\tt pos}={y}^{i}$, and $y^i_{\tt neg}=\widetilde{y}^{i,*}$. 

Using this preference dataset $\mathcal{D}_{\text{pref}}$, we further fine-tune $\pi_{\tt user}$ with the following DPO loss:
\begin{equation}
    \mathcal{L}_{\text{dpo}} =\underset{(x, y^{\text{pos}}, y^{\text{neg}})\in \mathcal{D}_{\text{pref}}}{\mathbb{E}}\left[-\log \sigma \left( \beta \cdot r_{\text{dpo}} \right) \right],\label{eq:ours_dpo}
\end{equation}
\noindent where $r_{\text{dpo}}=r_{\tt user}(y^{\text{pos}})-r_{\tt user}(y^{\text{neg}})$, and $\sigma(\cdot)$ denotes the sigmoid function.
Optimizing this loss encourages the personalized model $\pi_{\tt user}$ to assign higher reward to user-aligned responses compared to generic ones.
This better modeling of implicit user reward further improves the effectiveness of reward-guided decoding through \name{}.

\section{Experiments}

In this section, we design our experiments to investigate the following questions:
\begin{itemize}[leftmargin=5.5mm,topsep=10pt]
    \vspace{-0.1in}
    \item[$\circ$] Does \name{} yield better personalization than existing baselines?  (Table~\ref{tab:all_tasks_transposed})
    \vspace{-0.1in}
    \item[$\circ$] Is \name{} applicable to models of varying architectures and parameter scales? (Table~\ref{table:diff_llms})
    \vspace{-0.1in}
    \item[$\circ$] How do different components in \name{} contribute to personalization performance? (Table~\ref{tab:ablation_abstract_headline})
    \vspace{-0.1in}
    \item[$\circ$] How sensitive is the performance of \name{} to different configuration settings? (Figure~\ref{fig:fig_analysis})
\end{itemize}

\begin{table*}[t]
\centering
\caption{\textbf{Main Results.} ROUGE-1 and ROUGE-L scores are reported for five tasks: Abstract Generation, Review Writing, and Topic Writing from LongLaMP; News Headline Generation and Scholarly Title Generation from LaMP. All experiments are conducted using \texttt{Mistral-7B-Instruct-v0.3}.}

\renewcommand{\arraystretch}{1.2}
\resizebox{\textwidth}{!}{%
\begin{tabular}{l|cc|cc|cc|cc|cc}
\toprule
\multirow{2}{*}{\textbf{Methods}}  
& \multicolumn{2}{c|}{\textbf{Abstract Generation}} 
& \multicolumn{2}{c|}{\textbf{Review Writing}} 
& \multicolumn{2}{c|}{\textbf{Topic Writing}} 
& \multicolumn{2}{c|}{\textbf{News Headline}} 
& \multicolumn{2}{c}{\textbf{Scholarly Title}} \\
\cmidrule(r){2-3} \cmidrule(r){4-5} \cmidrule(r){6-7} \cmidrule(r){8-9} \cmidrule(r){10-11}
& ROUGE-1 & ROUGE-L 
& ROUGE-1 & ROUGE-L 
& ROUGE-1 & ROUGE-L 
& ROUGE-1 & ROUGE-L
& ROUGE-1 & ROUGE-L \\
\midrule
Base          
& 0.341 & 0.186     
& 0.287 & 0.126 
& 0.246 & 0.105  
& 0.119 & 0.105
& 0.409 & 0.324 \\
RAG          
& 0.347 & 0.205
& 0.272 & 0.128 
& 0.243 & 0.115
& 0.141 & 0.124
& 0.425 & 0.347 \\
PAG          
& 0.344 & 0.186   
& 0.256 & 0.125
& 0.262 & 0.107   
& 0.118 & 0.102
& 0.372 & 0.289 \\
TAM           
& 0.357 & 0.204    
& 0.289 & 0.122
& 0.253 & 0.107    
& 0.200 & 0.179 
& 0.514 & 0.456 \\
OPPU          
& 0.378 & 0.218    
& 0.319 & 0.134
& 0.278 & 0.112    
& 0.203 & 0.182 
& 0.510 & 0.454 \\ \midrule

\cc CoPE (Ours)    
& \cc \textbf{0.392} & \cc \textbf{0.239}    
& \cc \textbf{0.335} & \cc \textbf{0.146} 
& \cc \textbf{0.281} & \cc \textbf{0.120}    
& \cc \textbf{0.205} & \cc \textbf{0.184} 
& \cc \textbf{0.519} & \cc \textbf{0.461} \\
\bottomrule
\end{tabular}%
}
\label{tab:all_tasks_transposed}
\end{table*}

\subsection{Setups}

\paragraph{Datasets and metrics.} 
We evaluate the effectiveness of \name{} primarily on personalized text generation tasks from the Large Language Model Personalization (LaMP) \citep{salemi2024lamp} and LongLaMP \citep{kumar2024longlamp} benchmarks, which represent the most practical and impactful use cases of LLM personalization.  
In particular, we consider the following five tasks: News Headline Generation (LaMP 4), Scholarly Title Generation (LaMP 5), Abstract Generation (LongLaMP 2), Review Writing  (LongLaMP 3), and  Topic Writing (LongLaMP 4).\footnote{See behind rationale for this choice in Appendix \ref{app:datasets}.}
For evaluation, we mainly report ROUGE-1 and ROUGE-L scores across all tasks, which serve as standard evalaution metrics to measure the content relevance and structural similarity between the generated and ground-truth texts.\looseness=-1

\paragraph{Baselines.}  
We compare \name{} against several baselines to generate personalized responses from LLMs as follows: (1) \textit{Base} – generation using a vanilla model without any supervised fine-tuning; (2) \textit{RAG} \citep{lewis2021retrievalaugmentedgenerationknowledgeintensivenlp} – a retrieval-augmented generation method that directly injects user-related histories into the prompt without additional training; (3) \textit{PAG} \citep{richardson2023integratingsummarizationretrievalenhanced} – a prompt-augmented generation approach that additionally incorporates user profiles to the prompt; (4) \textit{TAM} \citep{tan2024democratizing} – generation with a task-adapted model trained on data from users excluding the test user, allowing familiarity with the task but lacking personalization; (5) \textit{OPPU} \citep{tan2024democratizing} – generation with a personalized model equipped with user-specific adapters trained via simple supervised fine-tuning on user data.

\paragraph{Implementation details.}

For methods that include a training step (TAM, OPPU, \name{}), all models are trained using AdamW \citep{loshchilov2019decoupledweightdecayregularization} with a weight decay of 0.01. 
Linear learning rate decay was used with a warm-up ratio of 0.1. 
The batch size for the initial training of the task-adapted model is set to 8, while subsequent training stages use 4 to better capture the style of each user.
Supervised training is conducted for 2 epochs with a learning rate of 1e-4 for LongLaMP and 1e-5 for LaMP. 
Subsequently, DPO training uses a 5e-6 learning rate for 1 epoch on LongLaMP and 2 epochs on LaMP.
Also, we note that OPPU is continuously applied after TAM, following \citet{tan2024democratizing}.
Similar to this, the proposed DPO step (Eq.~\ref{eq:ours_dpo}) is applied after OPPU (see Figure \ref{fig:cope_overview}).

All of the experiments are conducted using \texttt{Mistral-7B-Instruct-v0.3},\footnote{\url{https://huggingface.co/mistralai/Mistral-7B-Instruct-v0.3}} except for those reported in Table \ref{table:diff_llms}. 
Greedy decoding is used to eliminate randomness, except for negative sample generation. 
In this case, we use \texttt{vLLM} \citep{kwon2023efficientmemorymanagementlarge} with a temperature of 1.0 for faster decoding, generate 
\( K = 3 \)  candidates using the task-adapted model, and select the final negative using the reward function (Eq.~\ref{eq:reward_neg}).
For DPO training \citep{rafailov2023dpo}, we set coefficient for KL regularization \( \beta = 3.0 \) for LaMP tasks and \( \beta = 0.05 \) for LongLaMP tasks. At this point, we treat the task-adapted model as the base model $\pi_{\tt base}$ and the DPO-trained model as the user model $\pi_{\tt user}$ in Eq.~\ref{eq:reward}.
To implement the proposed reward-guided decoding (Eq.~\ref{eq:cope_decode}), we adopt the contrastive decoding \citep{Li2023contrastive}, 
with a plausibility threshold of \( \tau = 0.1 \) for both LaMP and LongLaMP tasks. 
The contrastive weight \( \alpha\) is set to 0.3 for LaMP and 0.1 for LongLaMP tasks. 
We apply a repetition penalty of 1.0 for LaMP and 7.0 for LongLaMP, after observing that these values offered acceptable control over repetition in preliminary experiments. 

\subsection{Main results}
Table \ref{tab:all_tasks_transposed} summarizes the experimental results on five personalized open-ended text generation tasks. 
First, it is observed that the effectiveness of prompting-based methods is indeed limited.
In particular, RAG and PAG exhibit limited improvement compared to training-based approaches, and even they are sometimes worse than the Base method, which does not apply any personalization technique.
This observation validates the necessity for developing a training-based method like the proposed framework.
Next, the experimental results in Table \ref{tab:all_tasks_transposed} also demonstrate that \name{} consistently outperforms all baseline methods across all tasks and metrics. 
For instance, \name{} achieves an average relative improvement of 10.57\% in ROUGE-L compared to the task-adapted model, TAM. 
Notably, \name{} even outperforms a personalized model OPPU that relies solely on explicit user-specific fine-tuning, with average relative improvement of 5.67\% in ROUGE-L. 
These results highlight the effectiveness of our framework, which maximizes implicit reward signals to better align with user preferences.\looseness=-1

We further observe a task-specific trend across benchmarks. While RAG shows some effectiveness in LaMP tasks, its performance declines in the LongLaMP setting. 
For instance, RAG scores 5.23\% lower than Base in Review Writing (ROUGE-1) and 1.22\% lower in Topic Writing (ROUGE-1). 
This highlights the increased difficulty of LongLaMP tasks, where simple retrieval of user history is no longer sufficient. 
In contrast, \name{} remains effective even in this more demanding setting. 
In fact, \name{} demonstrates a significantly higher relative improvement in the more challenging LongLaMP setting—achieving a 16.33\% gain in ROUGE-L over the task-adapted model, compared to just 1.95\% in LaMP. 
This suggests that LongLaMP tasks may offer greater room for personalization gains when properly modeled and carefully optimized.
We also note that the tasks in LongLaMP tend to involve more subjective or user-specific expression, making them especially well-suited for personalized generation when guided by an effective framework like \name{}.

\subsection{Additional analyses}

Here, we provide additional analyses of \name{} with the experiments on Abstract Generation from LongLaMP and News Headline Generation from LaMP. 
More analyses are in Appendix \ref{app:more_quant}. 

\begin{table}[t]
    \caption{\textbf{Compatibility of \name{}.} ROUGE-L scores on the Abstract Generation task across different LLMs.}
    \vspace{-0.1in}
    \begin{center}
\resizebox{1.0\linewidth}{!}{
    \begin{tabular}{c|ccc}
        \toprule
        Methods & LLaMA 3.1-8B & Gemma 3-4B & Qwen 2.5-1.5B \\ \midrule
        Base & 0.172 & 0.135 & 0.130 \\  
        RAG & 0.183 & 0.170 & 0.128 \\   
        PAG & 0.183 & 0.169 & 0.130 \\   
        TAM & 0.198 & 0.181 & 0.150 \\
        OPPU & 0.202 & 0.194 & 0.163 \\ \midrule
        \cc \name{} (Ours) & \cc \textbf{0.261} & \cc \textbf{0.237} & \cc \textbf{0.233} \\ \bottomrule
    \end{tabular}
    }
    \end{center}
    \vspace{-3.5mm}
    \label{table:diff_llms}
\end{table}

\paragraph{Generalization to various LLMs.} 
In this section, we explore the applicability of \name{} to various LLMs and sizes. Results are presented in Table \ref{table:diff_llms}.
The experimental results validate that \name{} generalizes well across a diverse range of LLMs, including LLaMA-3.1-8B-Instruct \citep{grattafiori2024llama}, Gemma-3-4B-it \citep{team2025gemma}, and Qwen2.5-1.5B-Instruct \citep{qwen2025qwen25technicalreport}. Compared to TAM, \name{} significantly improves ROUGE-L by 31.8\% on LLaMA-3.1-8B , 30.9\% on Gemma-3-4B-it, and 55.3\% on Qwen2.5-1.5B. 
Similarly, compared to OPPU, \name{} achieves a relative improvement of 29.2\% on LLaMA-3.1-8B, 22.2\% on Gemma-3-4B-it, and 42.9\% on Qwen2.5-1.5B. 
These consistent improvements suggest that \name{} does not simply rely on a specific environment setting. 
Instead, our framework is generalizable and flexible with respect to model architecture and parameter scale. 
This makes \name{} a broadly applicable framework for deployment across diverse LLMs.

\begin{figure*}[ht]
    \centering
    \begin{subfigure}{0.315\textwidth}
        \includegraphics[width=\linewidth]{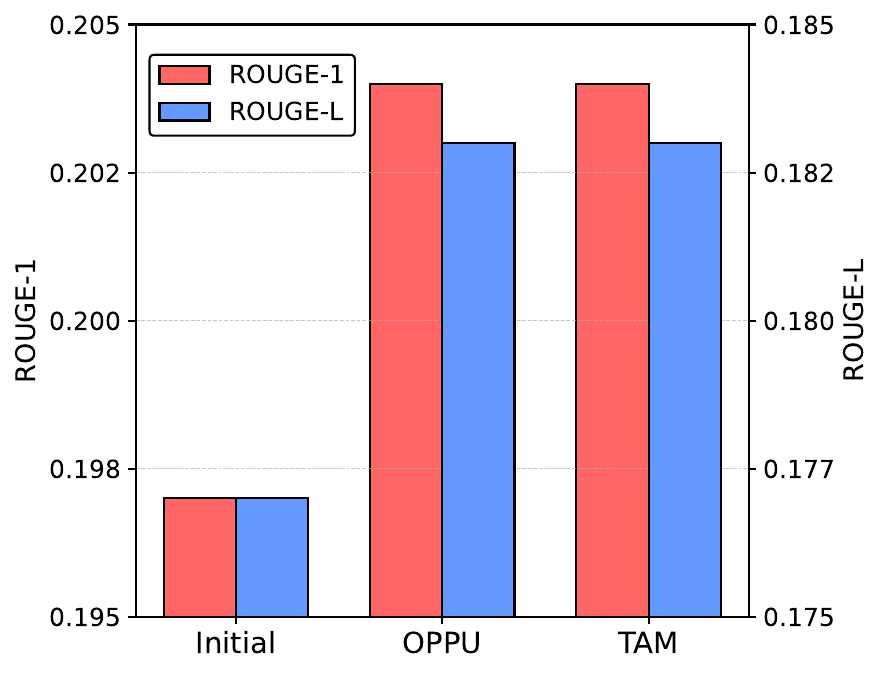}
        \caption{Choice of base model $\pi_{\tt base}$}
        \label{fig:fig2a}
    \end{subfigure}
    \hfill
    \begin{subfigure}{0.315\textwidth}
        \includegraphics[width=\linewidth]{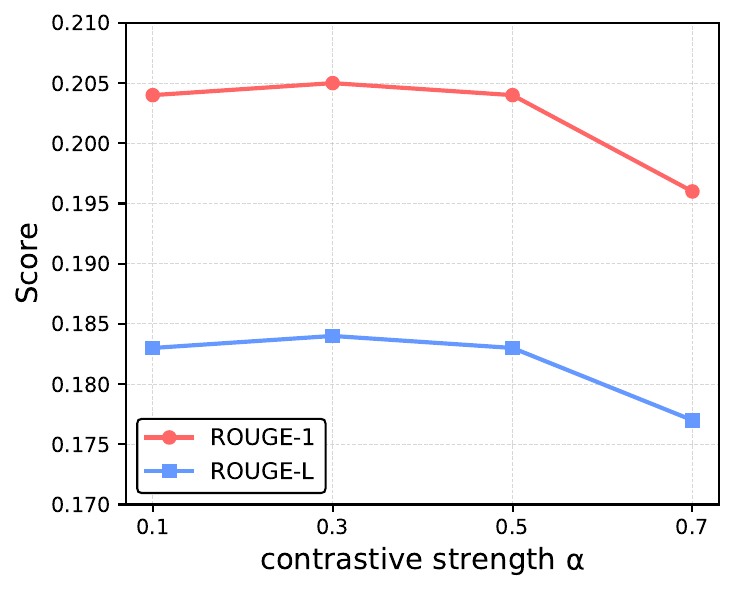}
        \caption{Contrasting strength $\alpha$}
        \label{fig:fig2b}
    \end{subfigure}
    \hfill
    \begin{subfigure}{0.315\textwidth}
        \includegraphics[width=\linewidth]{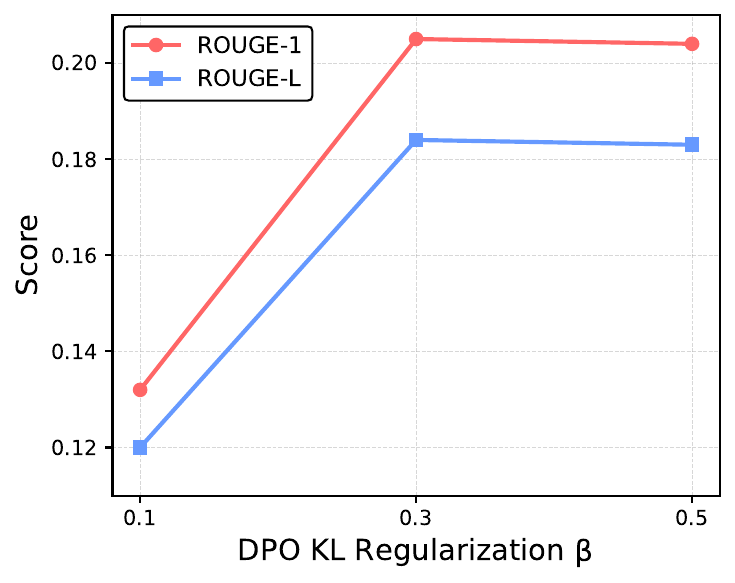}
        \caption{DPO KL regularization $\beta$}
        \label{fig:fig2c}
    \end{subfigure}

    \vspace{-0.1in}
    \caption{
        \textbf{Different hyperparameters.}
        (a) Performance variation by base model choice.
        (b) Effect of contrastive strength $\alpha$.
        (c) Effect of KL regularization $\beta$ in DPO.
        ROUGE-1 and ROUGE-L scores are reported.
    }
    \label{fig:fig_analysis}
\end{figure*}

\begin{figure*}[t]
    \centering
    \hspace*{0cm} 
    \includegraphics[
        width=1.0\textwidth,
        trim=0 150 0 100,
        clip
    ]{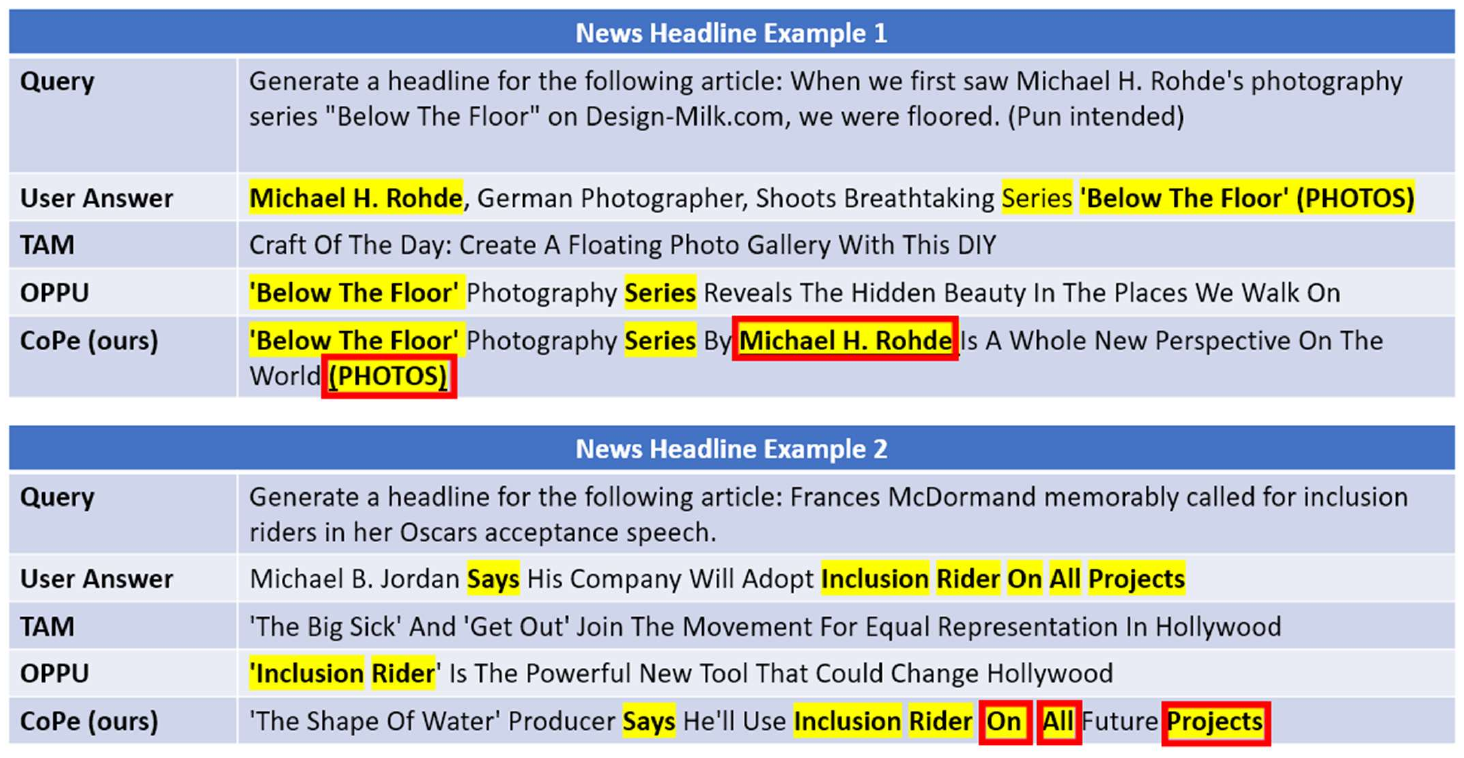}
    \caption{\textbf{A qualitative example of \name{} on the News Headline Generation task (LaMP 4).} The output of \name{} contains more words that align with the user gold response compared to TAM and OPPU. Words overlapping with the user’s answer are highlighted, and tokens that \name{} uniquely emphasizes for personalization, which are not captured by other baselines, are {\color{red}\textbf{boxed}}. More  qualitative examples from other tasks are provided in Appendix \ref{app:more_qual}.}

    \label{fig:qual_ex}
    \vspace{-3mm}
\end{figure*}

\paragraph{Ablation study.} We now proceed to validate the individual components of \name{}. 
To assess the contribution of each component to overall performance of \name{}, we perform a detailed ablation study.
\begin{table}[t]
\centering
\caption{\textbf{Ablation study.} 
The effects of contrastive decoding (CD) and direct preference optimization (DPO).}
\renewcommand{\arraystretch}{1.2}
\resizebox{1.0\linewidth}{!}{
\begin{tabular}{c|cc|cc|cc}
\toprule
 & \multirow{2}{*}{CD}  & \multirow{2}{*}{DPO}  
 & \multicolumn{2}{c|}{\textbf{Abstract Generation}} 
 & \multicolumn{2}{c}{\textbf{News Headline}} \\
 & & & \multicolumn{1}{c}{ROUGE-1} & \multicolumn{1}{c|}{ROUGE-L} 
   & \multicolumn{1}{c}{ROUGE-1} & \multicolumn{1}{c}{ROUGE-L} \\
\midrule
OPPU & {\color{red}\xmark} & {\color{red}\xmark} 
     & 0.378 & 0.218 
     & 0.203 & 0.181 \\
     & {\color{green}\cmark} & {\color{red}\xmark} 
     & 0.385 & 0.232 
     & 0.204 & 0.183 \\
     & {\color{red}\xmark} & {\color{green}\cmark} 
     & 0.386 & 0.230 
     & 0.203 & 0.182 \\
\midrule
\name{} (Ours) & {\color{green}\cmark} & {\color{green}\cmark} 
     & 0.392 & 0.239 
     & 0.205 & 0.184 \\
\bottomrule
\end{tabular}
}
\label{tab:ablation_abstract_headline}
\vspace{-0.1in}
\end{table}

For this analysis, we primarily conducted experiments on Abstract Generation and News Headline Generation tasks, serving as representative tasks for LongLaMP and LaMP, respectively.
The results are presented in Table \ref{tab:ablation_abstract_headline}. 
Here, it is observed that adding each component progressively improves the performance. 
Comparing with the OPPU baseline, applying only contrastive decoding increases the scores in both tasks, as it encourages the model to generate outputs that are more distinguishable from less preferred candidates. 
Meanwhile, in the training side, introducing only preference-aligned training also improves the performance of the model, as it guides the model to internalize user preferences by learning to favor higher-quality responses over inferior ones during fine-tuning.\looseness=-1

Finally, when combining these components to formulate an implicit reward maximization objective both during training and decoding, we observe the highest performance. 
These results indicate that each component independently contributes to performance improvements, and their integration yields the most substantial gains across tasks. 
This is because both components work synergistically to align model outputs with implicit user preferences: training encourages the model to internalize preference signals through comparisons between better and worse responses, while decoding promotes outputs that more closely reflect these learned preferences at inference time. 
Together, they implicitly guide the model to maximize a user-aligned reward signal, even in the absence of explicit supervision from an external model.\looseness=-1

\paragraph{Sensitivity of \name{}.}
Figure \ref{fig:fig_analysis} presents a sensitivity analysis of key components in the proposed framework. 
In this section, we conduct experiments on the News Headline Generation task, chosen for its shorter runtime, to explore the behavior of \name{} under different settings. \looseness=-1

We begin by examining the choice of base model for contrastive decoding (\textit{i.e.}, $\pi_{\tt base}$ to calculate likelihood for the denominator in Eq.~\ref{eq:reward}). 
We first note that TAM is originally used as the base model in \name{}, as it yields better understanding of the target task.
To investigate this, we performed experiments by varying the base models from TAM to init (\textit{i.e.}, initial Mistral model) and OPPU (\textit{i.e.}, after adaption to user and before DPO).
The results are presented in Figure \ref{fig:fig2a}, and one can verify that the current design choice is the best and using init is the worst. 
The findings suggest that using either OPPU or TAM as the base model yields the best performance. 
We hypothesize that these models help isolate and downweigh non-personalized features, allowing user-specific characteristics to be more prominently reflected.\looseness=-1

Next, we analyze the sensitivity of \name{} to two key hyperparameters: the contrastive strength ($\alpha$) and the KL regularization coefficient ($\beta$) in preference-aligned training. 
These two hyperparameters are crucial in the decoding and training components of our framework, respectively. 
Figure \ref{fig:fig2b} shows the effect of varying the contrastive strength $\alpha$ under fixed $\beta=3.0$.
We observe that \name{} performs reliably across a range of $\alpha$ values, with a slight peak around $\alpha=0.3$. 
While stronger contrastive signals may lead to marginal decreases in output quality, the overall performance remains consistently stable, which demonstrates the robustness of \name{} to decoding-time variations.\looseness=-1

Figure \ref{fig:fig2c} illustrates the impact of varying the KL regularization coefficient $\beta$ during training. 
As $\beta$ increases from 0.1 to 0.3, both ROUGE-1 and ROUGE-L scores improve, after which performance growth starts to hinder. 
This suggests that \name{} benefits from moderate regularization while remaining resilient to further increases. 
These results indicate that \name{} performs consistently well across a range of configurations, underscoring its robustness and reliability without signs of overfitting to specific hyperparameter values.\looseness=-1


\section{Conclusion}
\vspace{-0.5em}
In this work, we propose \name{}, the first decoding-based framework for personalizing LLMs.
Specifically, \name{} is a reward-guided decoding approach that maximizes implicit rewards of each user, thereby enhancing personalization without requiring external reward models.
Our comprehensive experiments show that \name{} consistently outperforms various baselines across multiple tasks and also is well-generalized to various types and scales of LLMs.
Consequently, these results demonstrate that it is not only effective but also a practical framework for decoding-time personalization.

\section*{Limitations}
While \name{} shows consistent improvements in personalized generation, it uses a fixed set of hyperparameters (\textit{e.g.}, learning rate, batch size, LoRA rank) for all users, regardless of dataset size or characteristics.
This uniform setting may be suboptimal when data varies in volume or domain.Future work should explore adaptive strategies that adjust hyperparameters to user-specific profiles.
In addition, we focus only on LoRA as the PEFT method, but different PEFT approaches \citep{li2021prefix, liu2022few} are also considerable. Since our approach does not depend on a particular method and most PEFT variants share architectural constraints with LoRA, we expect that \name{} is also easily deployed for these approaches. \looseness=-1
\section*{Ethics Statement}
We investigate LLM adaptation to individual users using PEFT methods such as LoRA. To ensure user privacy, our approach neither stores nor exposes raw user data and updates only a small set of task- and user-specific parameters.
In addition, all negative samples for preference optimization are synthetically generated from a base model, rather than extracted from real user outputs. 
Although we do not explicitly assess membership inference risks, the use of PEFT and synthetic negatives may provide stronger privacy protection than full-model fine-tuning.
All datasets and models used in this study are publicly available and used in line with their designated purposes. An AI assistant (ChatGPT) was used to refine the manuscript writing. \looseness=-1
\vspace{1.2em}
\section*{Acknowledgments}
This research was supported in part by Institute for Information \& communications Technology Planning \& Evaluation (IITP) grant funded by the Korea government (MSIT) (No.RS-2020-II201361, Artificial Intelligence Graduate School Program (Yonsei University); No. RS-2025-02215344, Development of AI Technology with Robust and Flexible Resilience Against Risk Factors).\looseness=-1
\vspace{-0.3em}

\bibliography{custom}

\begin{thebibliography}{48}
\providecommand{\natexlab}[1]{#1}

\bibitem[{Achiam et~al.(2023)Achiam, Adler, Agarwal, Ahmad, Akkaya, Aleman, Almeida, Altenschmidt, Altman, Anadkat et~al.}]{achiam2023gpt}
Josh Achiam, Steven Adler, Sandhini Agarwal, Lama Ahmad, Ilge Akkaya, Florencia~Leoni Aleman, Diogo Almeida, Janko Altenschmidt, Sam Altman, Shyamal Anadkat, et~al. 2023.
\newblock Gpt-4 technical report.
\newblock \emph{arXiv preprint arXiv:2303.08774}.

\bibitem[{Anthropic(2024)}]{claude3}
Anthropic. 2024.
\newblock Introducing the next generation of claude.
\newblock \emph{\url{https://www.anthropic.com/news/claude-3-family }}.

\bibitem[{Bradley and Terry(1952)}]{bradley1952rank}
Ralph~Allan Bradley and Milton~E Terry. 1952.
\newblock Rank analysis of incomplete block designs: I. the method of paired comparisons.
\newblock \emph{Biometrika}, 39(3/4):324--345.

\bibitem[{Chen et~al.(2025)Chen, Liu, Du, Pang, Liu, Sinha, Varakantham, and Lin}]{chen2025bootstrapping}
Changyu Chen, Zichen Liu, Chao Du, Tianyu Pang, Qian Liu, Arunesh Sinha, Pradeep Varakantham, and Min Lin. 2025.
\newblock Bootstrapping language models with dpo implicit rewards.
\newblock In \emph{International Conference on Learning Representations (ICLR)}.

\bibitem[{Chiang et~al.(2023)Chiang, Li, Lin, Sheng, Wu, Zhang, Zheng, Zhuang, Zhuang, Gonzalez, and et~al.}]{chiang2023vicuna}
Wei-Lin Chiang, Zhuohan Li, Zi~Lin, Ying Sheng, Zhanghao Wu, Hao Zhang, Lianmin Zheng, Siyuan Zhuang, Yonghao Zhuang, Joseph~E. Gonzalez, and et~al. 2023.
\newblock Vicuna: An open-source chatbot impressing gpt-4 with 90\%* chatgpt quality.
\newblock \url{https://vicuna.lmsys.org}.
\newblock Accessed 14 April 2023.

\bibitem[{Cui et~al.(2025)Cui, Yuan, Wang, Wang, Li, He, Fan, Yu, Xu, Chen et~al.}]{cui2025process}
Ganqu Cui, Lifan Yuan, Zefan Wang, Hanbin Wang, Wendi Li, Bingxiang He, Yuchen Fan, Tianyu Yu, Qixin Xu, Weize Chen, et~al. 2025.
\newblock Process reinforcement through implicit rewards.
\newblock \emph{arXiv preprint arXiv:2502.01456}.

\bibitem[{de~Masson~d'Autume et~al.(2019)de~Masson~d'Autume, Ruder, Kong, and Yogatama}]{dautume2019episodicmemorylifelonglanguage}
Cyprien de~Masson~d'Autume, Sebastian Ruder, Lingpeng Kong, and Dani Yogatama. 2019.
\newblock \href {https://arxiv.org/abs/1906.01076} {Episodic memory in lifelong language learning}.
\newblock \emph{Preprint}, arXiv:1906.01076.

\bibitem[{Deng and Raffel(2023)}]{deng2023reward}
Haikang Deng and Colin Raffel. 2023.
\newblock Reward-augmented decoding: Efficient controlled text generation with a unidirectional reward model.
\newblock \emph{arXiv preprint arXiv:2310.09520}.

\bibitem[{Grattafiori et~al.(2024)Grattafiori, Dubey, Jauhri, Pandey, Kadian, Al-Dahle, Letman, Mathur, Schelten, Vaughan et~al.}]{grattafiori2024llama}
Aaron Grattafiori, Abhimanyu Dubey, Abhinav Jauhri, Abhinav Pandey, Abhishek Kadian, Ahmad Al-Dahle, Aiesha Letman, Akhil Mathur, Alan Schelten, Alex Vaughan, et~al. 2024.
\newblock The llama 3 herd of models.
\newblock \emph{arXiv preprint arXiv:2407.21783}.

\bibitem[{Gui et~al.(2024)Gui, G{\^a}rbacea, and Veitch}]{gui2024bonbon}
Lin Gui, Cristina G{\^a}rbacea, and Victor Veitch. 2024.
\newblock Bonbon alignment for large language models and the sweetness of best-of-n sampling.
\newblock In \emph{Advances in Neural Information Processing Systems (NeurIPS)}.

\bibitem[{Hu et~al.(2021)Hu, Shen, Wallis, Allen-Zhu, Li, Wang, Wang, and Chen}]{hu2021loralowrankadaptationlarge}
Edward~J. Hu, Yelong Shen, Phillip Wallis, Zeyuan Allen-Zhu, Yuanzhi Li, Shean Wang, Lu~Wang, and Weizhu Chen. 2021.
\newblock \href {https://arxiv.org/abs/2106.09685} {Lora: Low-rank adaptation of large language models}.
\newblock \emph{Preprint}, arXiv:2106.09685.

\bibitem[{Hwang et~al.(2023)Hwang, Majumder, and Tandon}]{hwang2023aligning}
EunJeong Hwang, Bodhisattwa~Prasad Majumder, and Niket Tandon. 2023.
\newblock Aligning language models to user opinions.
\newblock In \emph{Conference on Empirical Methods in Natural Language Processing (EMNLP)}.

\bibitem[{Khanov et~al.(2024)Khanov, Burapacheep, and Li}]{khanov2024args}
Maxim Khanov, Jirayu Burapacheep, and Yixuan Li. 2024.
\newblock Args: Alignment as reward-guided search.
\newblock In \emph{International Conference on Learning Representations (ICLR)}.

\bibitem[{Kim et~al.(2025{\natexlab{a}})Kim, Lee, Shin, and Kim}]{kim2025spread}
Dongyoung Kim, Kimin Lee, Jinwoo Shin, and Jaehyung Kim. 2025{\natexlab{a}}.
\newblock Spread preference annotation: Direct preference judgment for efficient llm alignment.
\newblock In \emph{International Conference on Learning Representations (ICLR)}.

\bibitem[{Kim and Yang(2025)}]{kim2025few}
Jaehyung Kim and Yiming Yang. 2025.
\newblock Few-shot personalization of llms with mis-aligned responses.
\newblock In \emph{Conference of the North American Chapter of the Association for Computational Linguistics: Human Language Technologies (NAACL-HLT)}.

\bibitem[{Kim et~al.(2025{\natexlab{b}})Kim, Shin, and Kim}]{kim2025personalized}
Kyuyoung Kim, Jinwoo Shin, and Jaehyung Kim. 2025{\natexlab{b}}.
\newblock Personalized language models via privacy-preserving evolutionary model merging.
\newblock \emph{arXiv preprint arXiv:2503.18008}.

\bibitem[{Kumar et~al.(2024)Kumar, Viswanathan, Yerra, Salemi, Rossi, Dernoncourt, Deilamsalehy, Chen, Zhang, Agarwal et~al.}]{kumar2024longlamp}
Ishita Kumar, Snigdha Viswanathan, Sushrita Yerra, Alireza Salemi, Ryan~A Rossi, Franck Dernoncourt, Hanieh Deilamsalehy, Xiang Chen, Ruiyi Zhang, Shubham Agarwal, et~al. 2024.
\newblock Longlamp: A benchmark for personalized long-form text generation.
\newblock \emph{arXiv preprint arXiv:2407.11016}.

\bibitem[{Kwon et~al.(2023)Kwon, Li, Zhuang, Sheng, Zheng, Yu, Gonzalez, Zhang, and Stoica}]{kwon2023efficientmemorymanagementlarge}
Woosuk Kwon, Zhuohan Li, Siyuan Zhuang, Ying Sheng, Lianmin Zheng, Cody~Hao Yu, Joseph~E. Gonzalez, Hao Zhang, and Ion Stoica. 2023.
\newblock \href {https://arxiv.org/abs/2309.06180} {Efficient memory management for large language model serving with pagedattention}.
\newblock \emph{Preprint}, arXiv:2309.06180.

\bibitem[{Leng et~al.(2023)Leng, Zhang, Chen, Li, Lu, Miao, and Bing}]{leng2023mitigatingobjecthallucinationslarge}
Sicong Leng, Hang Zhang, Guanzheng Chen, Xin Li, Shijian Lu, Chunyan Miao, and Lidong Bing. 2023.
\newblock \href {https://arxiv.org/abs/2311.16922} {Mitigating object hallucinations in large vision-language models through visual contrastive decoding}.
\newblock \emph{Preprint}, arXiv:2311.16922.

\bibitem[{Lewis et~al.(2021)Lewis, Perez, Piktus, Petroni, Karpukhin, Goyal, Küttler, Lewis, tau Yih, Rocktäschel, Riedel, and Kiela}]{lewis2021retrievalaugmentedgenerationknowledgeintensivenlp}
Patrick Lewis, Ethan Perez, Aleksandra Piktus, Fabio Petroni, Vladimir Karpukhin, Naman Goyal, Heinrich Küttler, Mike Lewis, Wen tau Yih, Tim Rocktäschel, Sebastian Riedel, and Douwe Kiela. 2021.
\newblock \href {https://arxiv.org/abs/2005.11401} {Retrieval-augmented generation for knowledge-intensive nlp tasks}.
\newblock \emph{Preprint}, arXiv:2005.11401.

\bibitem[{Li et~al.(2023)Li, Holtzman, Fried, Liang, Eisner, Hashimoto, Zettlemoyer, and Lewis}]{Li2023contrastive}
Xiang~Lisa Li, Ari Holtzman, Daniel Fried, Percy Liang, Jason Eisner, Tatsunori Hashimoto, Luke Zettlemoyer, and Mike Lewis. 2023.
\newblock \href {https://aclanthology.org/2023.acl-long.687/} {Contrastive decoding: Open-ended text generation as optimization}.
\newblock In \emph{Proceedings of the 61st Annual Meeting of the Association for Computational Linguistics (Volume 1: Long Papers)}. Association for Computational Linguistics.

\bibitem[{Li and Liang(2021)}]{li2021prefix}
Xiang~Lisa Li and Percy Liang. 2021.
\newblock Prefix-tuning: Optimizing continuous prompts for generation.
\newblock In \emph{Annual Meeting of the Association for Computational Linguistics (ACL)}.

\bibitem[{Lightman et~al.(2024)Lightman, Kosaraju, Burda, Edwards, Baker, Lee, Leike, Schulman, Sutskever, and Cobbe}]{lightman2024let}
Hunter Lightman, Vineet Kosaraju, Yura Burda, Harri Edwards, Bowen Baker, Teddy Lee, Jan Leike, John Schulman, Ilya Sutskever, and Karl Cobbe. 2024.
\newblock Let's verify step by step.
\newblock In \emph{International Conference on Learning Representations (ICLR)}.

\bibitem[{Liu et~al.(2022)Liu, Tam, Muqeeth, Mohta, Huang, Bansal, and Raffel}]{liu2022few}
Haokun Liu, Derek Tam, Mohammed Muqeeth, Jay Mohta, Tenghao Huang, Mohit Bansal, and Colin~A Raffel. 2022.
\newblock Few-shot parameter-efficient fine-tuning is better and cheaper than in-context learning.
\newblock In \emph{Advances in Neural Information Processing Systems (NeurIPS)}.

\bibitem[{Loshchilov and Hutter(2019)}]{loshchilov2019decoupledweightdecayregularization}
Ilya Loshchilov and Frank Hutter. 2019.
\newblock \href {https://arxiv.org/abs/1711.05101} {Decoupled weight decay regularization}.
\newblock \emph{Preprint}, arXiv:1711.05101.

\bibitem[{Madotto et~al.(2021)Madotto, Lin, Zhou, Moon, Crook, Liu, Yu, Cho, Fung, and Wang}]{madotto-etal-2021-continual}
Andrea Madotto, Zhaojiang Lin, Zhenpeng Zhou, Seungwhan Moon, Paul Crook, Bing Liu, Zhou Yu, Eunjoon Cho, Pascale Fung, and Zhiguang Wang. 2021.
\newblock \href {https://aclanthology.org/2021.emnlp-main.590/} {Continual learning in task-oriented dialogue systems}.
\newblock In \emph{Proceedings of the 2021 Conference on Empirical Methods in Natural Language Processing}, Online and Punta Cana, Dominican Republic. Association for Computational Linguistics.

\bibitem[{McCloskey and Cohen(1989)}]{MCCLOSKEY1989109}
Michael McCloskey and Neal~J. Cohen. 1989.
\newblock \href {https://doi.org/10.1016/S0079-7421(08)60536-8} {Catastrophic interference in connectionist networks: The sequential learning problem}.
\newblock volume~24 of \emph{Psychology of Learning and Motivation}, pages 109--165. Academic Press.

\bibitem[{Mysore et~al.(2024)Mysore, Lu, Wan, Yang, Sarrafzadeh, Menezes, Baghaee, Gonzalez, Neville, and Safavi}]{mysore2024pearlpersonalizinglargelanguage}
Sheshera Mysore, Zhuoran Lu, Mengting Wan, Longqi Yang, Bahareh Sarrafzadeh, Steve Menezes, Tina Baghaee, Emmanuel~Barajas Gonzalez, Jennifer Neville, and Tara Safavi. 2024.
\newblock \href {https://arxiv.org/abs/2311.09180} {Pearl: Personalizing large language model writing assistants with generation-calibrated retrievers}.
\newblock \emph{Preprint}, arXiv:2311.09180.

\bibitem[{O'Brien and Lewis(2023)}]{obrien2023contrastivedecodingimprovesreasoning}
Sean O'Brien and Mike Lewis. 2023.
\newblock \href {https://arxiv.org/abs/2309.09117} {Contrastive decoding improves reasoning in large language models}.
\newblock \emph{Preprint}, arXiv:2309.09117.

\bibitem[{Ouyang et~al.(2022)Ouyang, Wu, Jiang, Almeida, Wainwright, Mishkin, Zhang, Agarwal, Slama, Ray, Schulman, Hilton, Kelton, Miller, Simens, Askell, Welinder, Christiano, Leike, and Lowe}]{ouyang2022traininglanguagemodelsfollow}
Long Ouyang, Jeff Wu, Xu~Jiang, Diogo Almeida, Carroll~L. Wainwright, Pamela Mishkin, Chong Zhang, Sandhini Agarwal, Katarina Slama, Alex Ray, John Schulman, Jacob Hilton, Fraser Kelton, Luke Miller, Maddie Simens, Amanda Askell, Peter Welinder, Paul Christiano, Jan Leike, and Ryan Lowe. 2022.
\newblock \href {https://arxiv.org/abs/2203.02155} {Training language models to follow instructions with human feedback}.
\newblock \emph{Preprint}, arXiv:2203.02155.

\bibitem[{Peng et~al.(2024)Peng, Liu, Xu, Yang, Shao, and Wang}]{peng2024reviewllmharnessinglargelanguage}
Qiyao Peng, Hongtao Liu, Hongyan Xu, Qing Yang, Minglai Shao, and Wenjun Wang. 2024.
\newblock \href {https://arxiv.org/abs/2407.07487} {Review-llm: Harnessing large language models for personalized review generation}.
\newblock \emph{Preprint}, arXiv:2407.07487.

\bibitem[{Qwen et~al.(2025)Qwen, :, Yang, Yang, Zhang, Hui, Zheng, Yu, Li, Liu, Huang, Wei, Lin, Yang, Tu, Zhang, Yang, Yang, Zhou, Lin, Dang, Lu, Bao, Yang, Yu, Li, Xue, Zhang, Zhu, Men, Lin, Li, Tang, Xia, Ren, Ren, Fan, Su, Zhang, Wan, Liu, Cui, Zhang, and Qiu}]{qwen2025qwen25technicalreport}
Qwen, :, An~Yang, Baosong Yang, Beichen Zhang, Binyuan Hui, Bo~Zheng, Bowen Yu, Chengyuan Li, Dayiheng Liu, Fei Huang, Haoran Wei, Huan Lin, Jian Yang, Jianhong Tu, Jianwei Zhang, Jianxin Yang, Jiaxi Yang, Jingren Zhou, Junyang Lin, Kai Dang, Keming Lu, Keqin Bao, Kexin Yang, Le~Yu, Mei Li, Mingfeng Xue, Pei Zhang, Qin Zhu, Rui Men, Runji Lin, Tianhao Li, Tianyi Tang, Tingyu Xia, Xingzhang Ren, Xuancheng Ren, Yang Fan, Yang Su, Yichang Zhang, Yu~Wan, Yuqiong Liu, Zeyu Cui, Zhenru Zhang, and Zihan Qiu. 2025.
\newblock \href {https://arxiv.org/abs/2412.15115} {Qwen2.5 technical report}.
\newblock \emph{Preprint}, arXiv:2412.15115.

\bibitem[{Rafailov et~al.(2023)Rafailov, Sharma, Mitchell, Ermon, Manning, and Finn}]{rafailov2023dpo}
Rafael Rafailov, Archit Sharma, Eric Mitchell, Stefano Ermon, Christopher~D. Manning, and Chelsea Finn. 2023.
\newblock \href {https://arxiv.org/abs/2305.18290} {Direct preference optimization: Your language model is secretly a reward model}.
\newblock In \emph{Proceedings of the 37th Conference on Neural Information Processing Systems (NeurIPS)}.

\bibitem[{Richardson et~al.(2023)Richardson, Zhang, Gillespie, Kar, Singh, Raeesy, Khan, and Sethy}]{richardson2023integratingsummarizationretrievalenhanced}
Chris Richardson, Yao Zhang, Kellen Gillespie, Sudipta Kar, Arshdeep Singh, Zeynab Raeesy, Omar~Zia Khan, and Abhinav Sethy. 2023.
\newblock \href {https://arxiv.org/abs/2310.20081} {Integrating summarization and retrieval for enhanced personalization via large language models}.
\newblock \emph{Preprint}, arXiv:2310.20081.

\bibitem[{Robertson and Walker(1994)}]{robertson1994okapi}
Stephen Robertson and Steve Walker. 1994.
\newblock Some simple effective approximations to the 2-poisson model for probabilistic weighted retrieval.
\newblock In \emph{Proceedings of the 17th annual international ACM SIGIR conference on Research and development in information retrieval}, pages 232--241. Springer.

\bibitem[{Salemi et~al.(2024)Salemi, Mysore, Bendersky, and Zamani}]{salemi2024lamp}
Alireza Salemi, Sheshera Mysore, Michael Bendersky, and Hamed Zamani. 2024.
\newblock Lamp: When large language models meet personalization.
\newblock In \emph{Annual Meeting of the Association for Computational Linguistics (ACL)}.

\bibitem[{Santurkar et~al.(2023)Santurkar, Durmus, Ladhak, Lee, Liang, and Hashimoto}]{santurkar2023whose}
Shibani Santurkar, Esin Durmus, Faisal Ladhak, Cinoo Lee, Percy Liang, and Tatsunori Hashimoto. 2023.
\newblock Whose opinions do language models reflect?
\newblock In \emph{Proceedings of the International Conference on Machine Learning (ICML)}.

\bibitem[{Shi et~al.(2023)Shi, Han, Lewis, Tsvetkov, Zettlemoyer, and tau Yih}]{shi2023trustingevidencehallucinatecontextaware}
Weijia Shi, Xiaochuang Han, Mike Lewis, Yulia Tsvetkov, Luke Zettlemoyer, and Scott~Wen tau Yih. 2023.
\newblock \href {https://arxiv.org/abs/2305.14739} {Trusting your evidence: Hallucinate less with context-aware decoding}.
\newblock \emph{Preprint}, arXiv:2305.14739.

\bibitem[{Tan et~al.(2024)Tan, Zeng, Tian, Liu, Yin, and Jiang}]{tan2024democratizing}
Zhaoxuan Tan, Qingkai Zeng, Yijun Tian, Zheyuan Liu, Bing Yin, and Meng Jiang. 2024.
\newblock Democratizing large language models via personalized parameter-efficient fine-tuning.
\newblock In \emph{Conference on Empirical Methods in Natural Language Processing (EMNLP)}.

\bibitem[{Tan et~al.(2025)Tan, Zeng, Tian, Liu, Yin, and Jiang}]{tan2025democratizinglargelanguagemodels}
Zhaoxuan Tan, Qingkai Zeng, Yijun Tian, Zheyuan Liu, Bing Yin, and Meng Jiang. 2025.
\newblock \href {https://arxiv.org/abs/2402.04401} {Democratizing large language models via personalized parameter-efficient fine-tuning}.
\newblock \emph{Preprint}, arXiv:2402.04401.

\bibitem[{Team et~al.(2023)Team, Anil, Borgeaud, Alayrac, Yu, Soricut, Schalkwyk, Dai, Hauth, Millican et~al.}]{team2023gemini}
Gemini Team, Rohan Anil, Sebastian Borgeaud, Jean-Baptiste Alayrac, Jiahui Yu, Radu Soricut, Johan Schalkwyk, Andrew~M Dai, Anja Hauth, Katie Millican, et~al. 2023.
\newblock Gemini: a family of highly capable multimodal models.
\newblock \emph{arXiv preprint arXiv:2312.11805}.

\bibitem[{Team et~al.(2025)Team, Kamath, Ferret, Pathak, Vieillard, Merhej, Perrin, Matejovicova, Ram{\'e}, Rivi{\`e}re et~al.}]{team2025gemma}
Gemma Team, Aishwarya Kamath, Johan Ferret, Shreya Pathak, Nino Vieillard, Ramona Merhej, Sarah Perrin, Tatiana Matejovicova, Alexandre Ram{\'e}, Morgane Rivi{\`e}re, et~al. 2025.
\newblock Gemma 3 technical report.
\newblock \emph{arXiv preprint arXiv:2503.19786}.

\bibitem[{Touvron et~al.(2023)Touvron, Martin, Stone, Albert, Almahairi, Babaei, Bashlykov, Batra, Bhargava, Bhosale, Bikel, Blecher, Ferrer, Chen, Cucurull, Esiobu, Fernandes, Fu, Fu, Fuller, Gao, Goswami, Goyal, Hartshorn, Hosseini, Hou, Inan, Kardas, Kerkez, Khabsa, Kloumann, Korenev, Koura, Lachaux, Lavril, Lee, Liskovich, Lu, Mao, Martinet, Mihaylov, Mishra, Molybog, Nie, Poulton, Reizenstein, Rungta, Saladi, Schelten, Silva, Smith, Subramanian, Tan, Tang, Taylor, Williams, Kuan, Xu, Yan, Zarov, Zhang, Fan, Kambadur, Narang, Rodriguez, Stojnic, Edunov, and Scialom}]{touvron2023llama2openfoundation}
Hugo Touvron, Louis Martin, Kevin Stone, Peter Albert, Amjad Almahairi, Yasmine Babaei, Nikolay Bashlykov, Soumya Batra, Prajjwal Bhargava, Shruti Bhosale, Dan Bikel, Lukas Blecher, Cristian~Canton Ferrer, Moya Chen, Guillem Cucurull, David Esiobu, Jude Fernandes, Jeremy Fu, Wenyin Fu, Brian Fuller, Cynthia Gao, Vedanuj Goswami, Naman Goyal, Anthony Hartshorn, Saghar Hosseini, Rui Hou, Hakan Inan, Marcin Kardas, Viktor Kerkez, Madian Khabsa, Isabel Kloumann, Artem Korenev, Punit~Singh Koura, Marie-Anne Lachaux, Thibaut Lavril, Jenya Lee, Diana Liskovich, Yinghai Lu, Yuning Mao, Xavier Martinet, Todor Mihaylov, Pushkar Mishra, Igor Molybog, Yixin Nie, Andrew Poulton, Jeremy Reizenstein, Rashi Rungta, Kalyan Saladi, Alan Schelten, Ruan Silva, Eric~Michael Smith, Ranjan Subramanian, Xiaoqing~Ellen Tan, Binh Tang, Ross Taylor, Adina Williams, Jian~Xiang Kuan, Puxin Xu, Zheng Yan, Iliyan Zarov, Yuchen Zhang, Angela Fan, Melanie Kambadur, Sharan Narang, Aurelien Rodriguez, Robert Stojnic, Sergey Edunov, and Thomas
  Scialom. 2023.
\newblock \href {https://arxiv.org/abs/2307.09288} {Llama 2: Open foundation and fine-tuned chat models}.
\newblock \emph{Preprint}, arXiv:2307.09288.

\bibitem[{Zhang et~al.(2024)Zhang, Sun, Chen, Lei, Abdul-Mageed, Wang, Jin, Park, Yao, and Long}]{zhang2024sparpersonalizedcontentbasedrecommendation}
Chiyu Zhang, Yifei Sun, Jun Chen, Jie Lei, Muhammad Abdul-Mageed, Sinong Wang, Rong Jin, Sem Park, Ning Yao, and Bo~Long. 2024.
\newblock \href {https://arxiv.org/abs/2402.10555} {Spar: Personalized content-based recommendation via long engagement attention}.
\newblock \emph{Preprint}, arXiv:2402.10555.

\bibitem[{Zhang et~al.(2025{\natexlab{a}})Zhang, Kim, and Liu}]{zhang2025personalizedllmresponsegeneration}
Kai Zhang, Yejin Kim, and Xiaozhong Liu. 2025{\natexlab{a}}.
\newblock \href {https://arxiv.org/abs/2404.03565} {Personalized llm response generation with parameterized memory injection}.
\newblock \emph{Preprint}, arXiv:2404.03565.

\bibitem[{Zhang et~al.(2025{\natexlab{b}})Zhang, Rossi, Kveton, Shao, Yang, Zamani, Dernoncourt, Barrow, Yu, Kim, Zhang, Gu, Derr, Chen, Wu, Chen, Wang, Mitra, Lipka, Ahmed, and Wang}]{zhang2025personalizationlargelanguagemodels}
Zhehao Zhang, Ryan~A. Rossi, Branislav Kveton, Yijia Shao, Diyi Yang, Hamed Zamani, Franck Dernoncourt, Joe Barrow, Tong Yu, Sungchul Kim, Ruiyi Zhang, Jiuxiang Gu, Tyler Derr, Hongjie Chen, Junda Wu, Xiang Chen, Zichao Wang, Subrata Mitra, Nedim Lipka, Nesreen Ahmed, and Yu~Wang. 2025{\natexlab{b}}.
\newblock \href {https://arxiv.org/abs/2411.00027} {Personalization of large language models: A survey}.
\newblock \emph{Preprint}, arXiv:2411.00027.

\bibitem[{Zhao et~al.(2024)Zhao, Dang, and Grover}]{zhao2024group}
Siyan Zhao, John Dang, and Aditya Grover. 2024.
\newblock Group preference optimization: Few-shot alignment of large language models.
\newblock In \emph{International Conference on Learning Representations (ICLR)}.

\bibitem[{Zhuang et~al.(2024)Zhuang, Sun, Yu, Qiang, Wang, Zhang, and Dai}]{zhuang2024hydra}
Yuchen Zhuang, Haotian Sun, Yue Yu, Rushi Qiang, Qifan Wang, Chao Zhang, and Bo~Dai. 2024.
\newblock Hydra: Model factorization framework for black-box llm personalization.
\newblock In \emph{Advances in Neural Information Processing Systems (NeurIPS)}.

\end{thebibliography}

\appendix

\newpage

\section{Datasets}\label{app:datasets}


\begin{table*}[t]
\centering
\small
\caption{\textbf{Dataset statistics.} 
Base LLM training corresponds to \textit{TAM}, and Personal PEFT training to OPPU.}
\renewcommand{\arraystretch}{1.1}
\resizebox{\textwidth}{!}{%
\begin{tabular}{l|ccc|ccc}
\toprule
\multirow{2}{*}{\textbf{Task}} & \multicolumn{3}{c|}{\textbf{Base LLM Training (TAM)}} & \multicolumn{3}{c}{\textbf{Personal PEFT Training (OPPU)}} \\
 & \#Train & $L_{\text{in}}$ & $L_{\text{out}}$ & \#Profile & $L_{\text{in}}$ & $L_{\text{out}}$ \\
\midrule
Abstract Generation & 31,808 & 70.4 $\pm$ 13.3 & 233.1 $\pm$ 117.5  & 1,296.7 $\pm$ 446.4 & 604.4 $\pm$ 142.7 & 210.5 $\pm$ 92.8 \\
Review Writing & 19,649 & 185.1 $\pm$ 109.0 & 407.2 $\pm$ 299.5  & 759.3 $\pm$ 324.2 & 1,143.0 $\pm$ 343.3 & 511.8 $\pm$ 294.2 \\
Topic Writing & 21,119 & 56.6 $\pm$ 54.8 & 358.3 $\pm$ 316.9  & 260.6 $\pm$ 314.0 & 759.8 $\pm$ 321.8 & 358.3 $\pm$ 255.4 \\
News Headline Generation & 7,275 & 53.6 $\pm$ 19.0 & 15.5 $\pm$ 6.0  & 270.1 $\pm$ 182.1 & 92.2 $\pm$ 11.3 & 18.6 $\pm$ 5.2 \\
Scholarly Title Generation & 16,076 & 230.6 $\pm$ 97.9 & 17.9 $\pm$ 6.1  & 444.0 $\pm$ 121.6 & 266.4 $\pm$ 85.9 & 16.4 $\pm$ 5.8 \\
\bottomrule
\end{tabular}
}
\end{table*}

For the experiments, we focus mainly on the text generation tasks provided in the LaMP \citep{salemi2024lamp} and LongLaMP \citep{kumar2024longlamp} benchmarks. Following these benchmarks, we use ROUGE-1 and ROUGE-L as metrics for evaluation. Detailed descriptions of each task are as follows.

\paragraph*{LaMP 4: News Headline Generation.}
This task evaluates the ability of a model to automatically generate headlines for given  news articles, conditioned on an author profile containing historical article-title pairs, thereby capturing distinctive stylistic patterns in journalistic writing.

\paragraph*{LaMP 5: Scholarly Title Generation.} This task assesses the capacity of a model to generate appropriate titles for scholarly article abstracts conditioned on an author profile of historical article-title pairs, reflecting distinct academic writing style.

\paragraph*{LongLaMP 2: Abstract Generation.} This task focuses on evaluating the proficiency of a model in generating scientific abstracts given paper titles and keywords by leveraging an author profile of previous publications to emulate characteristic academic writing style and domain-specific terminology

\paragraph*{LongLaMP 3: Review Writing.} This task tests the ability of a model to automatically generate comprehensive product reviews based on product specifications and user experiences, conditioned on a user profile of review history to reflect distinctive evaluative style and subjective perspective.

\paragraph*{LongLaMP 4: Topic Writing.} This task evaluates the capability of a model to generate Reddit post content based on post summaries while maintaining the unique writing style of individual users, requiring the generation of content from a given summary conditioned on a user profile containing their historical Reddit posts.\looseness=-1

\vspace{2mm}
In LaMP, we only consider News Headline Generation (LaMP 4) and Scholarly Title Generation (LaMP 5) as they are only applicable generation tasks with proper labels; Citation Identification (LaMP 1), Movie Tagging (LaMP 2), and Product Rating (LaMP 3) are discriminative, while Email Subject Generation (LaMP 6) and Tweet Paraphrasing (LaMP 7) lack gold labels, so none of these are included. For LongLaMP, we only considered Abstract Generation (LongLaMP 2), Review Writing (LongLaMP 3), and Topic Writing (LongLaMP 3) because Email Completion (LongLaMP 1) relies on the Avocado Research Email Collection, a private dataset with restricted access. Overall, our task selection focuses on (1) accessible and (2) evaluable (3) text-generation tasks for assessing LLM personalization.
Throughout our framework, we follow the setup of an earlier work \citep{tan2024democratizing}: we use 100 users with the longest activity histories as the test set, and the remaining users to train the task-adapted base model. 

\section{Baselines Details}
Detailed explanations for each baseline are provided below. 
Black boxes indicate vanilla models and prompt-base baselines (\textit{i.e.}, training-free), while white boxes represent training-base ones.

\begin{itemize}[label=\ding{110}]  
    \item \textbf{Base model} refers to the generation with the original LLM without any task-specific fine-tuning or additional conditioning. It represents the vanilla, commonly used standard pre-trained model as released.\looseness=-1

    \item \textbf{RAG: Retrieval-Augmented Generation} \citep{lewis2021retrievalaugmentedgenerationknowledgeintensivenlp} is a method that retrieves user-related history records and directly incorporates them into the prompt. Following the setup in LaMP \citep{salemi2024lamp}, we retrieve the top-$k$ history records for each user. In our experiments, we set $k = 3$, meaning the three most relevant records are selected using BM25 \citep{robertson1994okapi}—a standard keyword-based retrieval method. We implement BM25 using the \texttt{rank\_bm25} library with \texttt{BM25Okapi}.

    \item \textbf{PAG: Profile-Augmented Generation} \citep{richardson2023integratingsummarizationretrievalenhanced} is a technique for personalizing LLM outputs by conditioning on structured user profiles. Following the prior work \citep{tan2024democratizing}, we generate user profiles using the \texttt{vicuna-7B} model \citep{chiang2023vicuna}, based on the past responses of a typical user. Each profile captures key stylistic characteristics, such as tone, lexical choices, and recurring templates. The model then uses these profiles as a guide to generate output that aligns closely with the user style.
\end{itemize}

\begin{itemize}[label=\ding{113}] 
    \item \textbf{TAM: Task Adapted Model} \citep{tan2024democratizing} is trained on data from users other than the selected 100 test users. The objective of this model is to adapt the base model to the task in a general manner via LoRA (Low-Rank Adaptation) \citep{hu2021loralowrankadaptationlarge}, enabling it to understand the task setup without being exposed to the specific styles of the target users. 

    \item \textbf{OPPU: One PEFT Per User Model} \citep{tan2024democratizing} is a baseline that fine-tunes the LoRA adapter from the TAM model on individual users. Specifically, the historical data of each user is used to fine-tune the LoRA adapter from the TAM model, resulting in 100 separate personalized adapters. Intuitively, each LoRA adapter is specialized to learn the unique style of a specific user.
\end{itemize}

\section{More Quantitative Results}\label{app:more_quant}
In this section, we provide more quantitative results. 
In Table \ref{table:rouge1_abs}, we present the results under various LLMs on Abstract Generation using ROUGE-1, instead of ROUGE-L in Table \ref{table:diff_llms}. 
One can verify that \name{} significantly improve ROUGE-1 as well.
Next, in Tables \ref{table:rouge1_news} and \ref{table:rougel_news}, we present the results on News Headline Generation using ROUGE-1 and ROUGE-L scores, respectively.
Here, it is observed that the proposed \name{} is continuously effective.
\begin{table}[t]
    \caption{\textbf{Compatibility of \name{}.} ROUGE-1 scores on the Abstract Generation task across different LLMs.}
    \vspace{-0.1in}
    \begin{center}
    \resizebox{1.0\linewidth}{!}{
        \begin{tabular}{c|ccc}
            \toprule
            Methods & LLaMA 3.1-8B & Gemma 3-4B & Qwen 2.5-1.5B \\ \midrule
            Base & 0.340 & 0.270 & 0.278 \\  
            RAG  & 0.330 & 0.295 & 0.240 \\   
            PAG  & 0.333 & 0.292 & 0.241 \\   
            TAM  & 0.355 & 0.326 & 0.298 \\
            OPPU & 0.363 & 0.347 & 0.304 \\ \midrule
            \cc \name{} (Ours) & \cc \textbf{0.417} & \cc \textbf{0.393} & \cc \textbf{0.384} \\ \bottomrule
        \end{tabular}
    }
    \end{center}
    \label{table:rouge1_abs}
    \vspace{-0.1in}
\end{table}

\begin{table}[t]
    \caption{\textbf{Compatibility of \name{}.} ROUGE-1 scores on the News Headline Generation task across different LLMs.}
    \vspace{-0.1in}
    \begin{center}
    \resizebox{1.0\linewidth}{!}{
        \begin{tabular}{c|ccc}
            \toprule
            Methods & LLaMA 3.1-8B & Gemma 3-4B & Qwen 2.5-1.5B \\ \midrule
            Base & 0.127 & 0.070 & 0.117 \\  
            RAG  & 0.146 & 0.098 & 0.136 \\   
            PAG  & 0.129 & 0.099 & 0.128 \\   
            TAM  & 0.188 & 0.161 & 0.142 \\
            OPPU & 0.191 & 0.164 & 0.143 \\ \midrule
            \cc \name{} (Ours) & \cc \textbf{0.211} & \cc \textbf{0.168} & \cc \textbf{0.147} \\ \bottomrule
        \end{tabular}
    }
    \end{center}
    \label{table:rouge1_news}
    \vspace{-0.1in}
\end{table}

\begin{table}[t]
    \caption{\textbf{Compatibility of \name{}.} ROUGE-L scores on the News Headline Generation task across different LLMs.}
    \vspace{-0.1in}
    \begin{center}
    \resizebox{1.0\linewidth}{!}{
        \begin{tabular}{c|ccc}
            \toprule
            Methods & LLaMA 3.1-8B & Gemma 3-4B & Qwen 2.5-1.5B \\ \midrule
            Base & 0.110 & 0.063 & 0.104 \\  
            RAG  & 0.129 & 0.089 & 0.121 \\   
            PAG  & 0.112 & 0.089 & 0.114 \\   
            TAM  & 0.169 & 0.144 & 0.127 \\
            OPPU & 0.171 & 0.147 & 0.127 \\ \midrule
            \cc \name{} (Ours) & \cc \textbf{0.190} & \cc \textbf{0.151} & \cc \textbf{0.131} \\ \bottomrule
        \end{tabular}
    }
    \end{center}
    \label{table:rougel_news}
    \vspace{-0.1in}
\end{table}

\section{Empirical Validation of Log-Likelihood Ratios as Implicit Rewards}
\label{app:implicit-reward-validation}

\begin{table}[t]
\centering
\caption{\textbf{Validation of implicit reward approximation.} 
Comparison of log-likelihood ratio scores from user-specific models (\texttt{score\_user}) and models trained on other users (\texttt{score\_others}).}
\label{tab:implicit-reward-validation}
\renewcommand{\arraystretch}{1.2}
\resizebox{0.8\linewidth}{!}{
\begin{tabular}{c|cc}
\toprule
\multirow{2}{*}{\textbf{User Index}} & \textbf{User Own}  & \textbf{Other 9 Users Avg}  \\
& (\texttt{score\_user}) & (\texttt{score\_others}) \\
\midrule
3   & 0.087 & 0.025 \\
13  & 0.096 & 0.028 \\
14  & 0.131 & 0.023 \\
17  & 0.083 & 0.023 \\
28  & 0.097 & 0.029 \\
31  & 0.090 & 0.026 \\
35  & 0.082 & 0.025 \\
81  & 0.099 & 0.030 \\
86  & 0.088 & 0.025 \\
94  & 0.093 & 0.026 \\
\midrule
\textbf{Avg} & \textbf{0.095} & \textbf{0.026} \\
\bottomrule
\end{tabular}
}
\end{table}

To empirically validate our use of log-likelihood ratios as implicit reward signals, we conducted an analysis comparing scores produced by user-specific models and those produced by models trained on other users. 
For example, given user 1, we compute the log-likelihood ratio score (Eq.~\ref{eq:reward}). on user 1’s training samples using two models: (1) the OPPU model of user 1 (\texttt{score\_user}), and (2) OPPU models trained on all other users (\texttt{score\_others}). 
Our core intention with this validation is to demonstrate that the log-likelihood ratio with personalized model meaningfully captures user-specific signals. 
In other words, if it truly reflects personalization, then a model fine-tuned for a given user should consistently assign higher scores to that user’s own data than models fine-tuned for different users. 
To ensure robustness, we repeated this comparison across 10 randomly selected users. 
For each user, we evaluated the scores on a randomly sampled 20\% subset of their profile history. The results are presented in Table~\ref{tab:implicit-reward-validation}.

By comparing \texttt{score\_user} and \texttt{score\_others} across users and averaging over profile data, we find that each user’s model assigns higher scores to their own data. This shows that log-likelihood ratios capture personalized signals, supporting their use as an implicit reward approximator.
\label{app:extendedablationexperiments}
\section{Extended Ablation Experiments}

\vspace{-0.2cm}
A full ablation table covering all tasks is presented in Table~\ref{tab:ablation}. Overall, our proposed framework CoPe continues to show a clear trend of performance improvement upon each component (CD and DPO).

\begin{table}[t]
\centering
\caption{\textbf{Ablation study.} 
The effects of contrastive decoding (CD) and direct preference optimization (DPO). R-1 and R-L represents Rouge-1 and Rouge-L scores respectively.}
\renewcommand{\arraystretch}{1.2}

\resizebox{1.0\linewidth}{!}{
\begin{tabular}{c|cc|cc|cc|cc}
\toprule
 & \multirow{2}{*}{CD}  & \multirow{2}{*}{DPO}  
 & \multicolumn{2}{c|}{\textbf{Abstract Gen.}} 
 & \multicolumn{2}{c|}{\textbf{Review Writing}} 
 & \multicolumn{2}{c}{\textbf{Topic Writing}} \\
 & & & \multicolumn{1}{c}{R-1} & \multicolumn{1}{c|}{R-L} 
   & \multicolumn{1}{c}{R-1} & \multicolumn{1}{c|}{R-L} 
   & \multicolumn{1}{c}{R-1} & \multicolumn{1}{c}{R-L} \\
\midrule
OPPU & {\color{red}\xmark} & {\color{red}\xmark} 
     & 0.378 & 0.218 
     & 0.319 & 0.134 
     & 0.278 & 0.112 \\
     & {\color{green}\cmark} & {\color{red}\xmark} 
     & 0.385 & 0.232 
     & \textbf{0.335} & 0.145 
     & \textbf{0.285} & \textbf{0.122} \\
     & {\color{red}\xmark} & {\color{green}\cmark} 
     & 0.386 & 0.230 
     & 0.323 & 0.138 
     & 0.280 & 0.114 \\
\midrule
\name{} (Ours) & {\color{green}\cmark} & {\color{green}\cmark} 
     & \textbf{0.392} & \textbf{0.239} 
     & \textbf{0.335} & \textbf{0.146} 
     & 0.281 & 0.120 \\
\bottomrule
\end{tabular}
}

\vspace{0.5em}

\resizebox{1.0\linewidth}{!}{
\begin{tabular}{c|cc|cc|cc|cc}
\toprule
 & \multirow{2}{*}{CD}  & \multirow{2}{*}{DPO}  
 & \multicolumn{2}{c|}{\textbf{News Headline}} 
 & \multicolumn{2}{c|}{\textbf{Scholarly Title}} 
 & \multicolumn{2}{c}{\textbf{Average}} \\
 & & & \multicolumn{1}{c}{R-1} & \multicolumn{1}{c|}{R-L} 
   & \multicolumn{1}{c}{R-1} & \multicolumn{1}{c|}{R-L} 
   & \multicolumn{1}{c}{R-1} & \multicolumn{1}{c}{R-L} \\
\midrule
OPPU & {\color{red}\xmark} & {\color{red}\xmark} 
     & 0.203 & 0.181 
     & 0.510 & 0.454 
     & 0.338 & 0.220 \\
     & {\color{green}\cmark} & {\color{red}\xmark} 
     & 0.204 & 0.183 
     & 0.514 & 0.456 
     & 0.345 & 0.227 \\
     & {\color{red}\xmark} & {\color{green}\cmark} 
     & 0.203 & 0.182 
     & 0.517 & 0.457 
     & 0.342 & 0.224 \\
\midrule
\name{} (Ours) & {\color{green}\cmark} & {\color{green}\cmark} 
     & \textbf{0.205} & \textbf{0.184} 
     & \textbf{0.519} & \textbf{0.461} 
     & \textbf{0.346} & \textbf{0.230} \\
\bottomrule
\end{tabular}
}

\label{tab:ablation}
\vspace{-0.1in}
\end{table}

On average across tasks, DPO improved ROUGE-1 and ROUGE-L by 1.09\% and 2.3\%, while CD yielded higher gains (2.13\% ROUGE-1, 5.02\% ROUGE-L). Their combination achieved the best performance, with 2.51\% and 5.79\% increases over OPPU. Although a few tasks (e.g., Topic Writing) showed weaker results under both methods, likely due to universal hyperparameters being suboptimal, the overall trend supports their combined effectiveness.

\begin{table*}[t]
\centering
\caption{\textbf{Perplexity analysis.} 
Mean and standard deviation (std) of perplexity measured with Mistral-7B-Instruct-v0.3 as the reference LM.}
\label{tab:perplexity}
\renewcommand{\arraystretch}{1.2}
\resizebox{0.9\textwidth}{!}{
\begin{tabular}{l|c|c|c|c|c|c|c}
\toprule
\textbf{Task} & \textbf{Metric} & \textbf{Gold Data} & \textbf{TAM} & \textbf{OPPU} & \textbf{OPPU+CD} & \textbf{OPPU+DPO} & \textbf{\name{}} \\
\midrule
\multirow{2}{*}{Abstract Generation} 
 & mean & 214.43 & 13.27 & 12.34 & 7.01 & 13.00 & 15.91 \\
 & std  & 396.39 & 10.11 & 14.75 & 2.91 & 12.12 & 21.91 \\
\midrule
\multirow{2}{*}{Topic Writing} 
 & mean & 363.10 & 26.32 & 13.41 & 16.86 & 15.80 & 17.79 \\
 & std  & 439.49 & 21.40 & 11.30 & 20.12 & 15.13 & 20.21 \\
\midrule
\multirow{2}{*}{Review Writing} 
 & mean & 780.07 & 17.13 & 20.69 & 15.78 & 20.86 & 19.01 \\
 & std  & 1233.23 & 11.28 & 18.47 & 12.73 & 15.26 & 13.90 \\
\midrule
\multirow{2}{*}{News Headline} 
 & mean & 41.79 & 28.71 & 25.86 & 28.25 & 27.20 & 27.32 \\
 & std  & 26.30 & 30.34 & 17.26 & 19.09 & 17.81 & 17.88 \\
\midrule
\multirow{2}{*}{Scholarly Title} 
 & mean & 79.79 & 55.65 & 47.37 & 47.37 & 67.27 & 70.48 \\
 & std  & 64.57 & 40.29 & 30.72 & 30.72 & 49.52 & 54.24 \\
\bottomrule
\end{tabular}
}
\vspace{-0.1in}
\end{table*}

\begin{figure*}[t]
    \centering
    \hspace*{0cm} 
    \includegraphics[
        width=1.0\textwidth,
        trim=0 250 0 200,
        clip
    ]{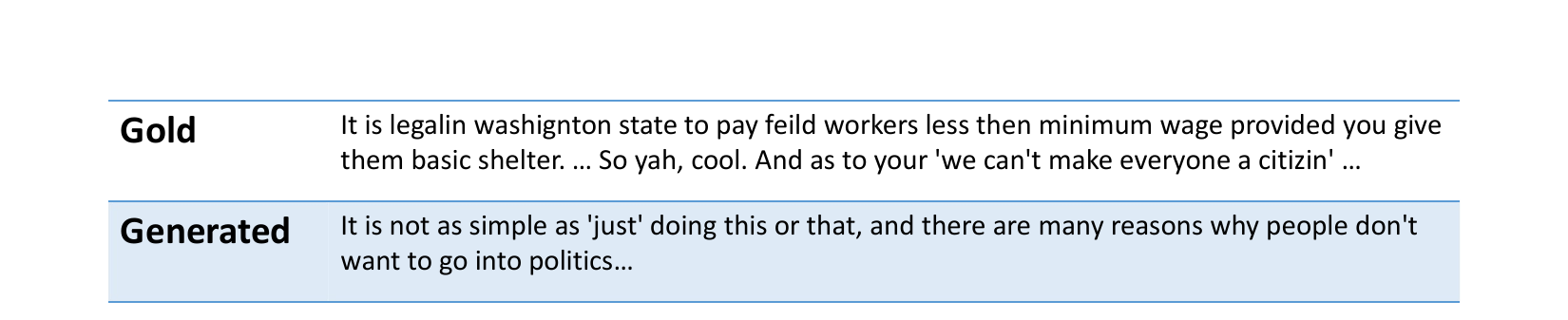}
    \caption{\textbf{Example illustrating perplexity differences between gold and generated text.} 
    A user's gold text often contains spelling variations, colloquial expressions, and conversational truncations, which are harder for a reference LM to predict, resulting in higher perplexity. 
    In contrast, model-generated text tends to be more regular and thus achieves lower perplexity.}
    \label{fig:perplexity-example}
    \vspace{-0.1in}
\end{figure*}

\section{Perplexity and Fluency Analysis}
\label{app:perplexity}

A common concern with decode-time manipulation methods such as contrastive decoding (CD) is whether they degrade fluency or coherence of the generated text. In this section, we explore the relationship between perplexity, our ratio-based reward proxy, and overall personalization quality.

Perplexity measures how well a reference model predicts a sequence in absolute terms. In contrast, the user/base probability ratio  (Eq.~\ref{eq:reward}) we employ is a relative measure that captures shifts in predicted token probabilities between a user-specific model and the base model. This relative shift serves as the personalization signal in our framework. Thus, while perplexity is informative about fluency under a particular reference model, it is not directly comparable to the ratio we use as implicit reward.

To empirically assess whether CD harms fluency, we measured the perplexity of all methods’ outputs using the same reference LM (\texttt{Mistral-7B-Instruct-v0.3}). Results are summarized in Table~\ref{tab:perplexity}. CD does not increase perplexity by more than $\pm$2.4 points compared to the base OPPU\_SFT, and in fact achieves the lowest perplexity on three of the five tasks (e.g., 7.01 vs.~12.34 on Abstract Generation). Additionally, standard deviations are small, which indicates stable behavior.

Furthermore, we also observe that human gold texts show higher perplexity than model outputs. This phenomenon is a natural outcome in personalization settings, as authentic user texts often include creative, informal, and unusual elements that are harder for a reference LM to predict.

Analytically, in  Figure~\ref{fig:perplexity-example}  the gold text contains spelling variations (``washignton'', ``feild'') ans colloquial phrases (``So yah, cool''), which raise perplexity under a reference LM, but this does not undermine our ratio-based reward, which relies on relative shifts between user-specific and base models rather than the absolute perplexity value itself.

\section{Robustness to User Heterogeneity}

While evaluating based on average performance across users is widely adopted, one may be concerned that the result can be over-estimated by few outlier users. 
Therefore, to test the robustness of \name{} under such heterogeneous user conditions, we compared it with TAM and the OPPU baseline at the instance level. Specifically, we measure how many instances are improved by our method compared to the baselines. 

\begin{table*}[t]
\centering
\caption{\textbf{Instance-level robustness comparison.}
“R-1 Improved (\%)” and “R-L Improved (\%)” denote the percentage of test instances for which the model outperforms TAM. 
The \textbf{Average} row reports a \emph{macro-average across the five tasks} (unweighted by instance counts); hence Total Instances is not applicable.}
\label{tab:user_robustness}
\renewcommand{\arraystretch}{1.2}

\resizebox{0.9\linewidth}{!}{
\begin{tabular}{l|c|cc|c c|c|c}
\toprule
\textbf{Task} & \textbf{Model} & \textbf{CD} & \textbf{DPO} & \textbf{R-1 Improved (\%)} & \textbf{R-L Improved (\%)} & \textbf{Total Instances} \\
\midrule
\multirow{4}{*}{Abstract Generation}
  & OPPU   & {\color{red}\xmark} & {\color{red}\xmark} & 50.35 & 52.45 & \multirow{4}{*}{143} \\
  & + CD   & {\color{green}\cmark} & {\color{red}\xmark} & 58.04 & 65.03 & \\
  & + DPO  & {\color{red}\xmark} & {\color{green}\cmark} & 56.64 & 66.43 & \\
  & \cc{\name{} (ours)} & \cc{{\color{green}\cmark}} & \cc{{\color{green}\cmark}} & \cc{\textbf{63.64}} & \cc{\textbf{76.22}} & \\
\midrule
\multirow{4}{*}{Review Writing}
  & OPPU   & {\color{red}\xmark} & {\color{red}\xmark} & 69.33 & 68.10 & \multirow{4}{*}{163} \\
  & + CD   & {\color{green}\cmark} & {\color{red}\xmark} & 75.46 & 75.46 & \\
  & + DPO  & {\color{red}\xmark} & {\color{green}\cmark} & 68.10 & 72.39 & \\
  & \cc{\name{} (ours)} & \cc{{\color{green}\cmark}} & \cc{{\color{green}\cmark}} & \cc{\textbf{77.91}} & \cc{\textbf{81.60}} & \\
\midrule
\multirow{4}{*}{Topic Writing}
  & OPPU   & {\color{red}\xmark} & {\color{red}\xmark} & 58.33 & 60.61 & \multirow{4}{*}{132} \\
  & + CD   & {\color{green}\cmark} & {\color{red}\xmark} & 56.06 & 65.15 & \\
  & + DPO  & {\color{red}\xmark} & {\color{green}\cmark} & \textbf{59.09} & 55.30 & \\
  & \cc{\name{} (ours)} & \cc{{\color{green}\cmark}} & \cc{{\color{green}\cmark}} & \cc{56.06} & \cc{\textbf{65.91}} & \\
\midrule
\multirow{4}{*}{News Headline}
  & OPPU   & {\color{red}\xmark} & {\color{red}\xmark} & 61.01 & 60.86 & \multirow{4}{*}{6725} \\
  & + CD   & {\color{green}\cmark} & {\color{red}\xmark} & 56.85 & 57.71 & \\
  & + DPO  & {\color{red}\xmark} & {\color{green}\cmark} & 61.32 & 61.56 & \\
  & \cc{\name{} (ours)} & \cc{{\color{green}\cmark}} & \cc{{\color{green}\cmark}} & \cc{\textbf{79.61}} & \cc{\textbf{80.24}} & \\
\midrule
\multirow{4}{*}{Scholarly Title}
  & OPPU   & {\color{red}\xmark} & {\color{red}\xmark} & 52.83 & 53.77 & \multirow{4}{*}{106} \\
  & + CD   & {\color{green}\cmark} & {\color{red}\xmark} & 52.83 & 53.77 & \\
  & + DPO  & {\color{red}\xmark} & {\color{green}\cmark} & \textbf{65.09} & \textbf{66.98} & \\
  & \cc{\name{} (ours)} & \cc{{\color{green}\cmark}} & \cc{{\color{green}\cmark}} & \cc{56.60} & \cc{53.77} & \\
\midrule
\multirow{4}{*}{\textbf{Average}}
  & OPPU   & {\color{red}\xmark} & {\color{red}\xmark} & 58.37 & 59.16 & \multirow{4}{*}{-} \\
  & + CD   & {\color{green}\cmark} & {\color{red}\xmark} & 59.85 & 63.43 & \\
  & + DPO  & {\color{red}\xmark} & {\color{green}\cmark} & 62.05 & 64.53 & \\
  & \cc{\name{} (ours)} & \cc{{\color{green}\cmark}} & \cc{{\color{green}\cmark}} & \cc{\textbf{66.77}} & \cc{\textbf{71.55}} & \\
\bottomrule
\end{tabular}
}

\vspace{-0.1in}
\end{table*}

\begin{table}[t]
\centering
\small
\caption{\textbf{Mean ROUGE and standard error (SE) across test samples.} 
SE values remain consistently small ($<0.009$), indicating stable results.}
\label{tab:se_results}
\resizebox{1.0\linewidth}{!}{\begin{tabular}{lcccc}
\toprule
Model & Mean R-1 $\uparrow$ & Mean R-L $\uparrow$ & SE R-1 $\downarrow$ & SE R-L $\downarrow$ \\
\midrule
TAM & 0.323 & 0.214 & 0.008 & 0.007 \\
OPPU & 0.338 & 0.220 & 0.009 & 0.007 \\
\cc{\name{} (ours)} & \cc{\textbf{0.346}} & \cc{\textbf{0.228}} & \cc{0.009} & \cc{0.007} \\
\bottomrule
\end{tabular}}

\end{table}

As shown in Table~\ref{tab:user_robustness}, \name{} outperformed TAM on 56\%–80\% of instances, while OPPU improved on only about 50-69\%. 
Although this analysis is conducted at the instance level, it provides strong evidence that even under a fixed hyper-parameter setting, \name{} generalizes robustly across heterogeneous writing styles, domains, and data sizes. 
We leave exploring efficient, light-weight user-specific hyper-parameter adaptation as promising future work to pursue.\looseness=-1

\section{Statistical Reliability of Results}

To assess the reliability and significance of the reported performance, 
we quantify the variability of ROUGE scores by measuring the standard error (SE) 
across test samples. Table~\ref{tab:se_results} shows the averaged results across datasets. 
Both ROUGE-1 and ROUGE-L exhibit very small SE values (below $\approx 0.009$), 
corresponding to only 2–3\% of the mean score and yielding a 95\% confidence interval 
within $\pm 0.02$ ROUGE points. 
This indicates that the observed improvements are unlikely to be due to random variation 
in the evaluation set.

\begin{table}[t]
\centering
\small
\caption{\textbf{Instance-level win rates against TAM.} Values denote the percentage of test instances where a model’s ROUGE is 
equal to or higher than TAM.}
\label{tab:winrate_results}
\begin{tabular}{lcc}
\toprule
Model & R-1 $\uparrow$ & R-L $\uparrow$ \\
\midrule
OPPU & 58.37 & 59.16 \\
\cc{\name{} (ours)} & \cc{\textbf{66.77}} & \cc{\textbf{71.55}} \\
\bottomrule
\end{tabular}
\end{table}

\vspace{0.1in}
In addition, to complement mean/SE figures with an instance-level view, we compute the percentage of test instances (IDs) where a model’s ROUGE score matches or exceeds TAM. As shown in Table~\ref{tab:winrate_results}, \name{} surpasses TAM on 66.77\% of instances for ROUGE-1 and 71.55\% for ROUGE-L, about 8 and 12 points above OPPU. Overall, the low SE values and high win-or-tie rates show that \name{}’s gains are statistically reliable and consistently realized at the instance level.

\section{Background for RLHF and DPO}\label{app:rlhf_dpo}

Let us denote LLM as $\pi_{\theta}$, which generates an output sequence (\textit{e.g.}, response) $y$ for a given input sequence (\textit{e.g.}, prompt) $x$, \textit{i.e.,} $y \sim \pi_{\theta}(\cdot|x)$.   
Then, the goal of LLM alignment is to make $\pi_{\theta}$ provide human-aligned responses to various input prompts.  
To this end, let assume that the preference dataset $\mathcal{D}=\{(x,y_l,y_w)\}$ is available which consists of the triplets of input prompt $x$, preferred response $y_w$, and dispreferred response $y_l$. 
Here, the preference labels were annotated by a ground truth annotator, that is usually a human expert. 

\noindent\paragraph{Reward modeling and RL fine-tuning.} 
Since a pairwise preference between $y_w$ and $y_l$ is hard to model directly, one of the common practices is introducing reward function $r(x,y)$ and modeling the preference based on this using the Bradley-Terry model \looseness=-1\citep{bradley1952rank}:
\begin{equation*}
p(y_{w} \succ y_{l} \mid x) = \frac{\exp\left(r(x, y_w)\right)}{\exp\left(r(x, y_w)\right) + \exp\left(r(x, y_l)\right)}.
\label{eq:bt_reward}
\end{equation*}

\noindent From this, one can introduce a parametrized reward model $r_{\phi}(x,y)$ by estimating its parameters with the maximum-likelihood objective:
\begin{equation*}
    \mathcal{L}_r= \underset{(x, y_w, y_l) \sim \mathcal{D}}{\mathbb{E}}\left[- \log \sigma \left( r_\phi(x, y_w) - r_\phi(x, y_l) \right) \right],
\end{equation*}
where $\sigma$ is a sigmoid function. 
After this reward modeling procedure, one could improve the alignment of LLM $\pi_{\theta}$ by optimizing it to maximize the reward from $r_\phi$.  
Here, KL-distance from the reference model $\pi_{\text{ref}}$ is incorporated as a regularization to prevent the reward over-optimization of $\pi_{\theta}$, with a hyper-parameter $\beta > 0$ \citep{ouyang2022traininglanguagemodelsfollow}:\footnote{$\pi_{\text{ref}}$ is usually initialized with supervised fine-tuned (SFT) LLM. 
Also, $\pi_{\theta}$ is initialized with $\pi_{\text{ref}}$.} 
\begin{align*}
\mathcal{L}_\text{RLHF} = 
& -\mathbb{E}_{y \sim \pi_{\theta},\, x \sim \rho} \left[ r_{\phi}(x, y) \right] \notag \\
& + \beta D_{\mathrm{KL}}\left( \pi_{\theta}(y|x) \| \pi_{\text{ref}}(y|x) \right).
\label{eq:rlhf}
\end{align*}

\noindent\paragraph{Direct preference optimization.} 
\citet{rafailov2023dpo} propose an alternative approach to align LLM $\pi_{\theta}$ with the preference dataset $\mathcal{D}$, which is called Direct Preference Optimization (DPO).  
DPO integrates a two-step alignment procedure with reward modeling and RL fine-tuning into a single unified fine-tuning procedure. 
Specifically, the optimal reward function is derived from the RLHF objective, with the target LLM $\pi_{\theta}$ and the reference model $\pi_{\text{ref}}$, which is often called implicit reward: 
\begin{equation*}\label{eq:orig_reward}
r(x, y) = \beta \log \frac{\pi_{\theta}(y \mid x)}{\pi_{\text{ref}}(y \mid x)} + \beta \log Z(x), 
\end{equation*}
where $Z(x) = \sum_y \pi_{\text{ref}}(y \mid x) \exp \left( \frac{1}{\beta} r(x, y) \right)$. 
Then, the preference between two responses could be measured using this reward derivation, and $\pi_{\theta}$ is optimized to maximize this preference of $y_w$ over $y_l$ using the preference dataset $\mathcal{D}$.
\begin{align*}
p_{\theta}(y_{w} \succ y_{l} \mid x) 
&= \sigma \bigg( \beta \log \frac{\pi_{\theta}(y_w \mid x)}{\pi_{\text{ref}}(y_w \mid x)} \notag \\
&\hspace{3.8em} - \beta \log \frac{\pi_{\theta}(y_l \mid x)}{\pi_{\text{ref}}(y_l \mid x)} \bigg), \label{eq:orig_pref}
\end{align*}
\begin{equation*}\label{eq:dpo_objective}
    \mathcal{L}_\text{DPO} = \mathbb{E}_{(x, y_w, y_l) \sim \mathcal{D}} \left[-\log p_{\theta}(y_{w} \succ y_{l} | x) \right].
\end{equation*}

\section{Prompts}
Below are prompts used in our experiments. Note that the text in $\texttt{\{BRACES\}}$ is a placeholder for user- and query-specific input.

\paragraph*{News Headline Generation}~\\
\textit{You are a news headline generator.\\  
Generate a headline for the following article. \\
\noindent article: $\texttt{\{ARTICLE\}}$  \\
\noindent headline:}

\paragraph*{Scholarly Title Generation}~\\
\textit{You are a scholarly title generator.\\  
Generate a title for the following abstract of a paper.
\noindent abstract: $\texttt{\{ABSTRACT\}}$  \\
\noindent title:}

\paragraph*{Abstract Generation}~\\
\textit{You are an abstract writer.\\  
Generate the review text written by a reviewer who has a given an overall rating of "$\texttt{\{RATING\}}$" for a product with description "$\texttt{\{PRODUCT\}}$". The summary of the review text is "$\texttt{\{SUMMARY\}}$".\\
\noindent Review:}

\paragraph*{Review Writing}~\\
\textit{You are a review writer.\\  
Generate an abstract for the title “$\texttt{\{TITLE\}}$”.\\
\noindent Abstract:}

\paragraph*{Topic Writing}~\\
\textit{You are a creative content generator for Reddit posts.\\  
Generate the content for a reddit post.\\
\noindent post: $\texttt{\{POST\}}$\\
\noindent content:}

\section{Chat Templates}
In this section, we provide the chat templates we applied for experiments. We also include the chat templates of other LLMs used to test the generalization of \name{}.

\vspace{1em}

\noindent\textbf{\texttt{Mistral-7B-Instruct-v0.3}}
\begin{lstlisting}[language=Python]
MISTRAL_CHAT_TEMPLATE = """
{% if messages[0]['role'] == 'system' %}
{% set loop_messages = messages[1:] %}
{% set system_message = messages[0]['content'].strip() + '\n' %}
{% else %}
{% set loop_messages = messages %}
{% set system_message = '' %}
{% endif %}
{% for message in loop_messages %}
    {% if loop.index0 == 0 %}
        {% set content = system_message + message['content'] %}
    {% else %}
        {% set content = message['content'] %}
    {% endif %}
    {% if message['role'] == 'user' %}
        {{ '[INST] ' + content.strip() + ' [/INST]' }}
    {% elif message['role'] == 'assistant' %}
        {{ ' ' + content.strip() + ' ' + eos_token }}
    {% endif %}
{% endfor %}
"""
\end{lstlisting}

\vspace{1em}
\noindent\textbf{\texttt{LLaMA-3.1-8B-Instruct}}

\vspace{0.3em}
\begin{lstlisting}[language=Python]
LLAMA_CHAT_TEMPLATE = """
 {{- bos_token }}
 {%- if messages[0]['role'] == 'system' %}
     {%- set system_message = messages[0]['content'].strip() %}
     {%- set loop_messages = messages[1:] %}
     {{- '<|start_header_id|>system<|end_header_id|>\\n\\n' + system_message + '<|eot_id|>' }}
 {%- else %}
     {%- set loop_messages = messages %}
 {%- endif %}
 {%- for message in loop_messages %}
     {%- if message['role'] == 'user' %}
         {{- '<|start_header_id|>user<|end_header_id|>\\n\\n' + message['content'].strip() + '<|eot_id|>' }}
     {%- elif message['role'] == 'assistant' %}
         {{- '<|start_header_id|>assistant<|end_header_id|>\\n\\n' + message['content'].strip() + '<|eot_id|>' }}
     {%- endif %}
 {%- endfor %}
 {%- if add_generation_prompt %}
     {{- '<|start_header_id|>assistant<|end_header_id|>\\n\\n' }}
 {%- endif %}"""
\end{lstlisting}

\vspace{1em}
\noindent\textbf{\texttt{GEMMA-3-4B-it}} 

\vspace{0.3em}
\begin{lstlisting}[language=Python]
GEMMA_CHAT_TEMPLATE = """
"{% set bos_token = '<bos>' %}
{% set eos_token = '<eos>' %}

{{ bos_token }}
{% if messages[0]['role'] == 'system' %}
  {{ 'System: ' + messages[0]['content'].strip() + '\n' }}
  {% set loop_messages = messages[1:] %}
{% else %}
  {% set loop_messages = messages %}
{% endif %}

{% for message in loop_messages %}
  {% if message['role'] == 'user' %}
    {{ 'User: ' + message['content'].strip() + '\n' }}
  {% elif message['role'] == 'assistant' %}
    {{ 'Assistant: ' + message['content'].strip() + eos_token + '\n' }}
  {% endif %}
{% endfor %}
{{ 'Assistant:' }}"
"""
\end{lstlisting}

\vspace{1em}
\noindent\textbf{\texttt{Qwen2.5-1.5B-Instruct}}

\vspace{0.3em}
\begin{lstlisting}[language=Python]
QWEN_CHAT_TEMPLATE = ''' {%- if messages[0]['role'] == 'system' %}
     {{- '<|im_start|>system\\n' + messages[0]['content'].strip() + '<|im_end|>\\n' }}
     {%- set loop_messages = messages[1:] %}
 {%- else %}
     {%- set loop_messages = messages %}
 {%- endif %}
 {%- for message in loop_messages %}
     {%- if message['role'] == 'user' %}
         {{- '<|im_start|>user\\n' + message['content'].strip() + '<|im_end|>\\n' }}
     {%- elif message['role'] == 'assistant' %}
         {{- '<|im_start|>assistant\\n' + message['content'].strip() + '<|im_end|>\\n' }}
     {%- endif %}
 {%- endfor %}
 {%- if add_generation_prompt %}
     {{- '<|im_start|>assistant\\n' }}
 {%- endif %}
'''
\end{lstlisting}

\section{More Qualitative Examples}\label{app:more_qual}

\begin{figure*}[t]
  \centering
  \includegraphics[width=\textwidth, trim=0 20 0 10, clip]{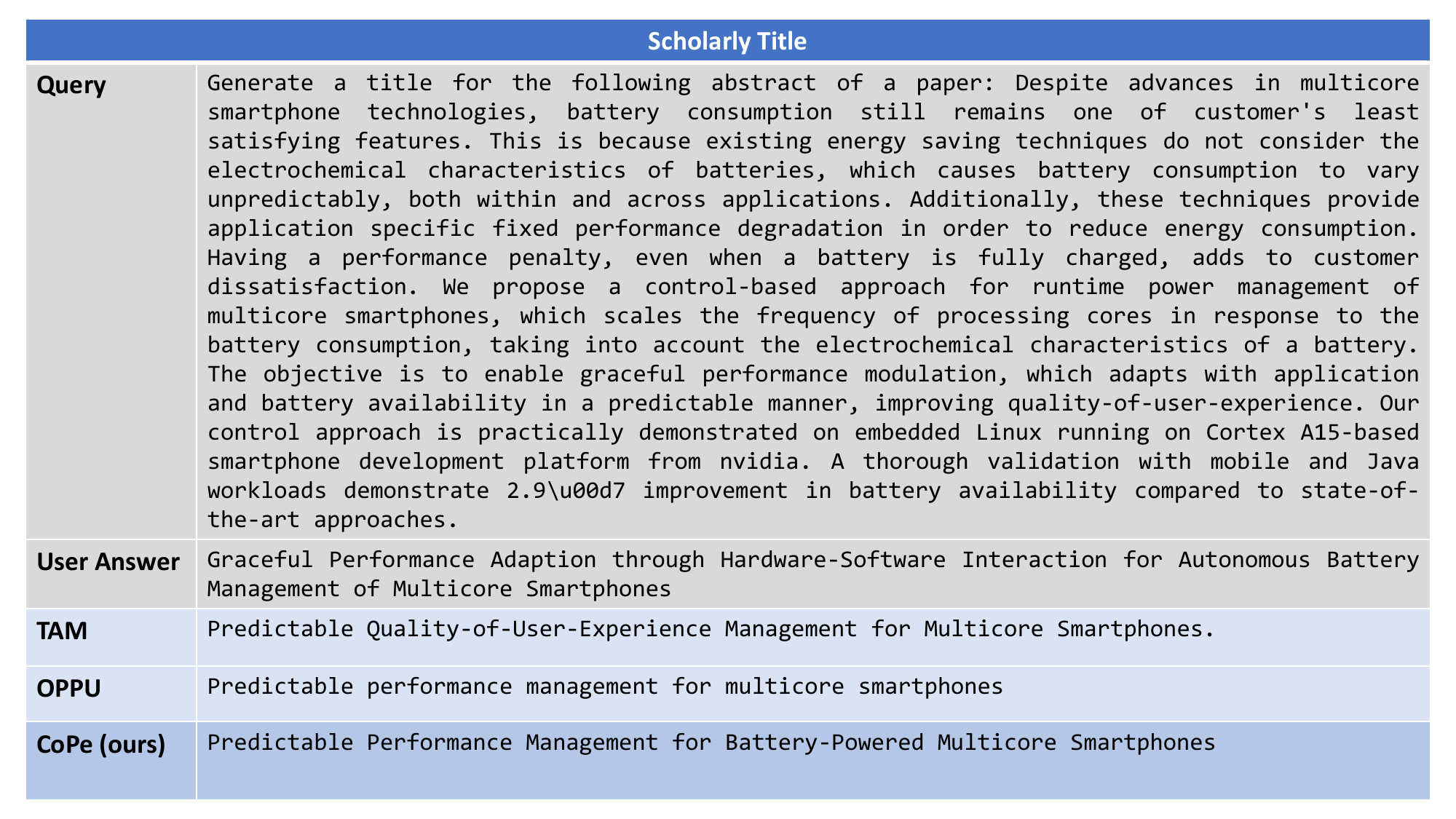}
  \caption{\textbf{Qualitative example for Scholarly Title Generation}}
  \label{fig:scholarly}
\end{figure*}

\begin{figure*}[t]
  \centering
  \includegraphics[width=\textwidth, trim=0 20 0 10, clip]{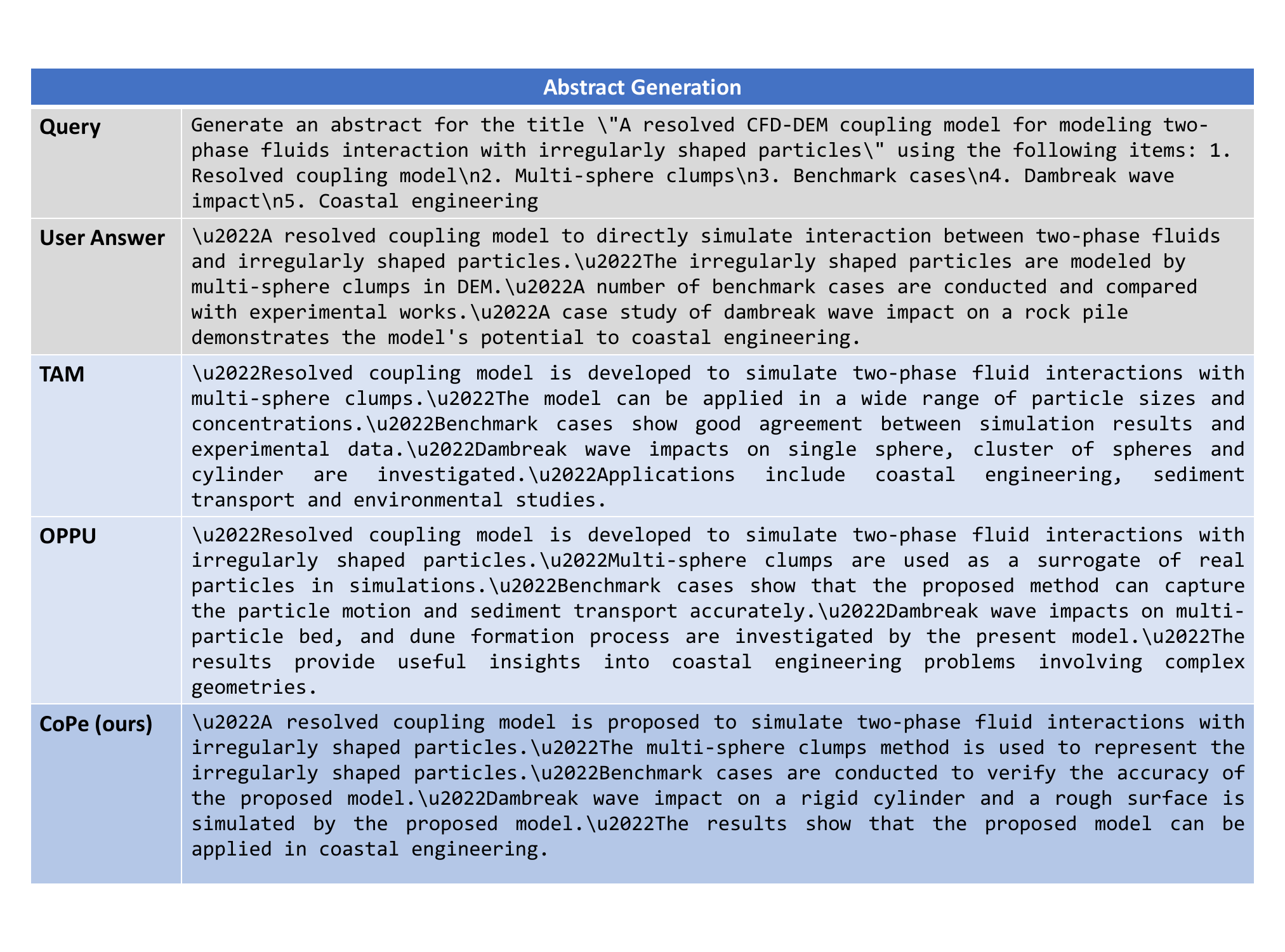}
  \caption{\textbf{Qualitative example for Abstract Generation}}
  \label{fig:abstract}
\end{figure*}

\begin{figure*}[t]
  \centering
  \includegraphics[width=\textwidth, trim=0 20 0 10, clip]{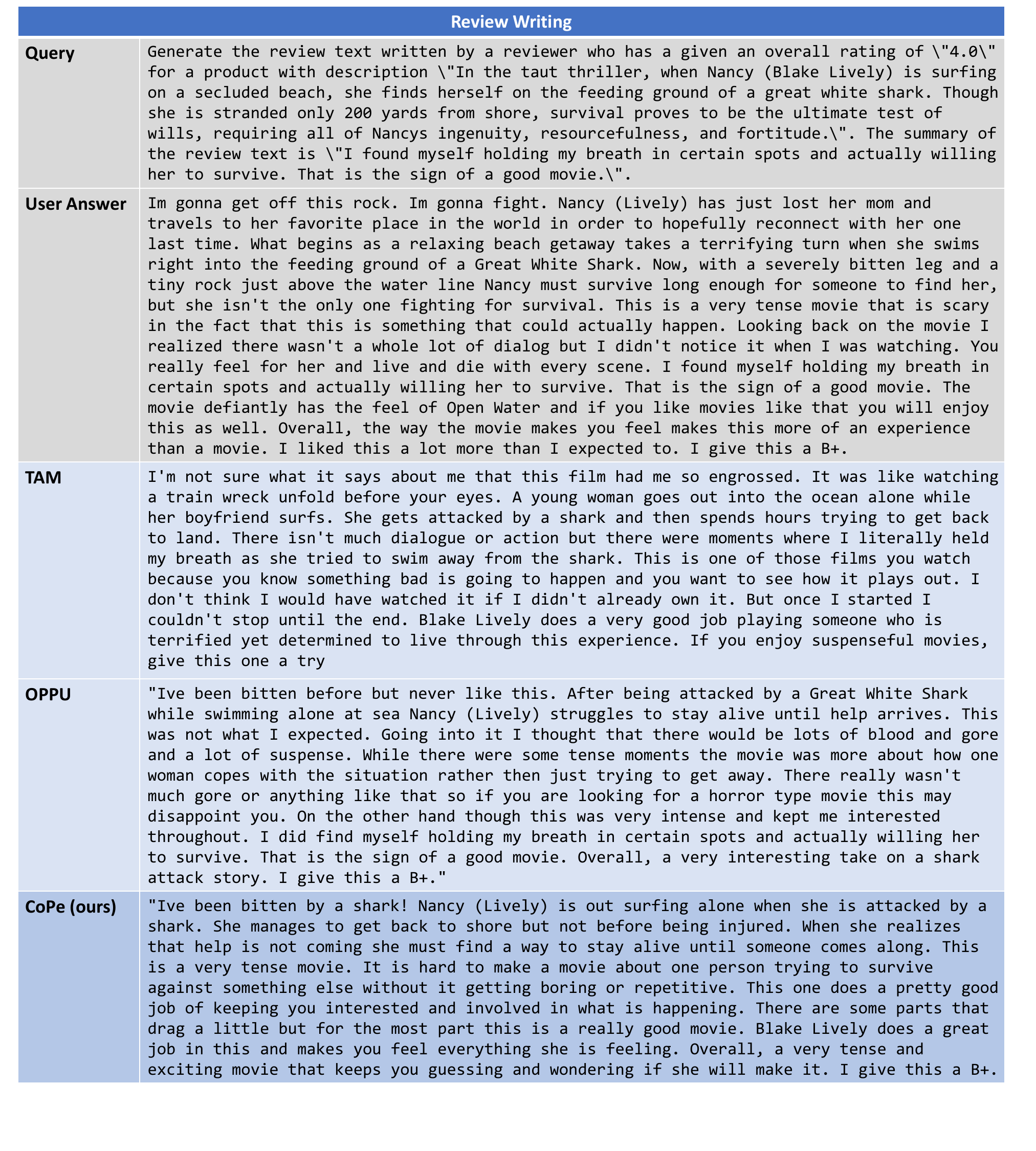}
  \caption{\textbf{Qualitative example for Review Writing}}
  \label{fig:review}
\end{figure*}

\begin{figure*}[t]
  \centering
  \includegraphics[width=\textwidth, trim=0 20 0 10, clip]{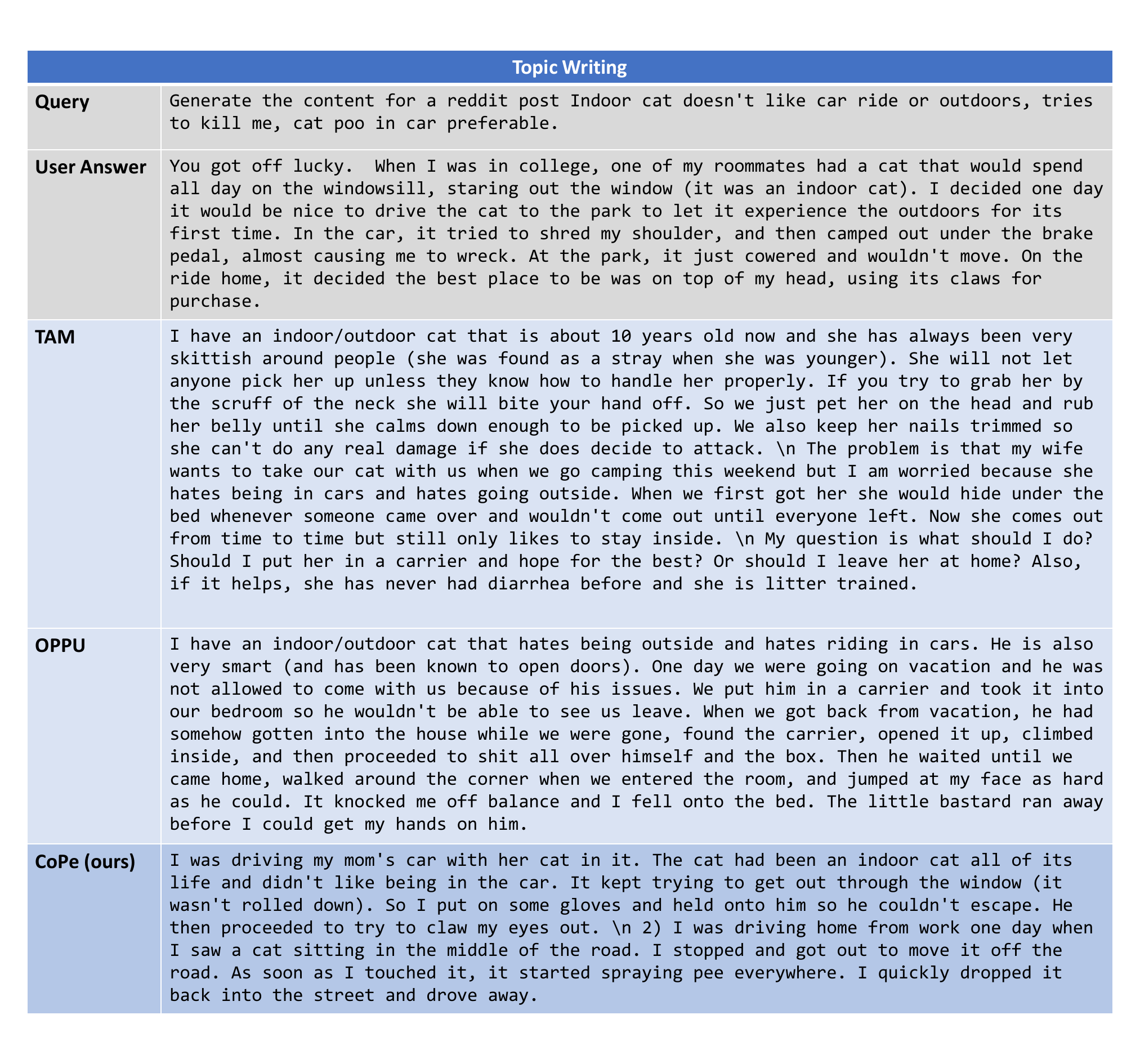}
  \caption{\textbf{Qualitative example for Topic Writing}}
  \label{fig:topic}
\end{figure*}

In this section, we present the additional qualitative examples similar to Figure \ref{fig:qual_ex}. Figures \ref{fig:scholarly}, \ref{fig:abstract}, \ref{fig:review}, and \ref{fig:topic} clearly show the advantages of \name{}, compared to the baseline methods.

\end{document}